\documentclass{article}

\usepackage{adjustbox}

\usepackage{tabularray}
\usepackage{lscape}
\usepackage{xcolor}

\usepackage{ulem}
\usepackage{rotating}  
\usepackage{bigstrut}  
\usepackage{float}

\usepackage{times}

\usepackage{latexsym}
\usepackage{graphicx}

\usepackage{booktabs}
\usepackage{comment}
\usepackage{multirow}
\usepackage{arabtex}
\usepackage{pifont}
\usepackage{utf8}
\usepackage{amsmath}
\usepackage{natbib}

\usepackage[T1]{fontenc}

\usepackage[utf8]{inputenc}

\usepackage{microtype}
\usepackage{tabularx}

\usepackage{graphicx}
\begin{document}

\title{Arabic Text Diacritization In The Age Of Transfer Learning: Token Classification Is All You Need} 

\author{
    Abderrahman Skiredj \\
    OCP Solutions and College of Computing \\
    Mohammed VI Polytechnic University \\
    \texttt{abderrahman.skiredj@ocpsolutions.ma}\\
    \&
    Ismail Berrada \\
    College of Computing \\
    Mohammed VI Polytechnic University \\
    \texttt{ismail.berrada@um6p.ma}
}

\author{
    Abderrahman Skiredj\thanks{OCP Solutions and College of Computing, Mohammed VI Polytechnic University; Email: \texttt{abderrahman.skiredj@ocpsolutions.ma}} \and 
    Ismail Berrada\thanks{College of Computing, Mohammed VI Polytechnic University; Email: \texttt{ismail.berrada@um6p.ma}}
}

\maketitle

\begin{abstract}
Automatic diacritization of Arabic text involves adding diacritical marks (diacritics) to the text. This task poses a significant challenge
with noteworthy implications for computational processing and comprehension. In this paper, we introduce PTCAD (Pre-FineTuned
Token Classification for Arabic Diacritization, a novel two-phase approach for the Arabic Text Diacritization task.  PTCAD comprises a pre-finetuning phase and a finetuning phase, treating Arabic Text Diacritization as a token classification task for pre-trained models. The effectiveness of PTCAD is demonstrated through evaluations on two benchmark datasets derived from the Tashkeela dataset, where it achieves state-of-the-art results, including a 20\% reduction in Word Error Rate (WER) compared to existing benchmarks and superior performance over GPT-4 in ATD tasks.
\end{abstract}

\maketitle

\section{Introduction}

Natural Language Processing (NLP) has experienced marked success in recent years, reshaping a myriad of applications such as text analysis \cite{textclassif}, machine translation \cite{attention}, Large Language Models (LLMs), and conversational models \cite{gpt4}. These strides have not only heightened the automation and effectiveness of textual data processes but have also played a pivotal role in fields like education \cite{education} and customer service \cite{custo} among others. While the majority of these developments have been in the context of the English language, given its prevalent usage and vast data availability, it is evident that unique linguistic characteristics inherent in other languages necessitate tailored approaches for comparable success.

In the realm of Arabic NLP, notable advancements have been achieved, encompassing dialect identification \citep{el2020weighted, el-mekki-etal-2021-bert}, sentiment analysis \citep{mahdaouy2021deep}, and domain adaptation \citep{el2021domain}. Yet, Arabic, a Semitic language, possesses linguistic nuances that diverge greatly from English. A prime example of this is the role of diacritics in Arabic \citep{survey}. These diminutive marks, placed above or below characters, are indispensable for clarifying meaning and pronunciation. In Modern Standard Arabic (MSA) \cite{msa}, the frequent omission of diacritics introduces a notable level of ambiguity in written form. Such omissions pose challenges to NLP tasks, highlighting the importance of Arabic Text Diacritization (ATD) as a crucial task in Arabic NLP. ATD is significant for several reasons:

\begin{itemize}
    \item Disambiguation. Arabic is typically written without diacritics, which are marks that indicate the vowels and pronunciation of words. The absence of diacritics can lead to ambiguity in the meaning of words, as many Arabic words share the same root consonants. Diacritization helps disambiguate these words, improving overall comprehension \citep{alhajj2004arabic,habash2007arabic}.
    \item NLP Applications. Many NLP applications, such as Text-To-Speech (TTS) generation, machine translation, reading comprehension, and Part-Of-Speech (POS) tagging, rely on accurately understanding the structure and meaning of text. Diacritization enhances the performance of these applications by providing more context and improving the accuracy of language processing \citep{hassan2010sakhr,abdul2012arabic}.
    \item Accessibility. Diacritization makes Arabic text more accessible to learners and individuals who are not native speakers. It aids in pronunciation and comprehension, making it easier for non-Arabic speakers to understand and learn the language \citep{taji2014impact}.
    \item Search and Information Retrieval. Diacritization can improve the accuracy of search and information retrieval systems. When diacritized, Arabic text becomes more searchable, helping users find relevant information more efficiently \citep{abdul2012arabic}.
    \item Preservation of Cultural and Linguistic Heritage. Diacritization contributes to the preservation of the richness and subtleties of the Arabic language. It helps maintain the integrity of classical texts and allows for a more accurate representation of the language's historical and cultural nuances \citep{taji2014impact}.
\end{itemize}

Thus, Arabic Text Diacritization is crucial for enhancing language processing applications, improving accessibility for learners, facilitating accurate information retrieval, and preserving the linguistic and cultural heritage of the Arabic language \citep{alhajj2004arabic,habash2007arabic,hassan2010sakhr,abdul2012arabic,taji2014impact}.

Prior work has explored various approaches to address the ATD task, spanning from rule-based methods to classical machine learning models, and contemporary deep learning architectures \citep{survey}, often integrated with pre-existing linguistic knowledge \citep{joint_diac_lemma}. Significant progress has notably been achieved in the diacritization of both Classical Arabic (CA) and MSA. For instance, in CA, \citet{multi_comp} attained a Diacritic Error Rate (DER) of 3.39\% and a Word Error Rate (WER) of 9.94\% on the cleaned version of the Tashkeela dataset, using a multi-layer Recurrent Neural Network (RNN) in conjunction with rule-based and statistical correction components. Similarly, in MSA, \citet{joint_diac_lemma} employed a joint modeling approach, achieving an impressive diacritization accuracy of 93.9\% in the Penn Arabic Treebank (PATB) dataset \cite{patb}. Despite these advancements, a significant challenge persists: the complexity and richness of the Arabic language necessitate a deep contextual understanding to achieve accurate diacritization. Furthermore, existing large language models, such as ChatGPT \citep{gpt4}, often struggle with handling the inherent ambiguity present in undiacritized text. This challenge becomes particularly evident when models are trained on data that is either inaccurate or inconsistent in terms of quality and standards.  Specifically, ChatGPT encounters difficulties in achieving precise diacritization and tends to demonstrate a lack of full comprehension when processing complex Arabic sentences devoid of diacritics.

To tackle these challenges, this paper introduces PTCAD (Pre-FineTuned Token Classification for Arabic Diacritization), a novel two-phase approach to ATD. The core idea of PTCAD revolves around framing the ATD task as a finetuning task for pre-trained BERT-like models, leveraging their robustness in encapsulating contextual information. Initially, PTCAD undergoes simultaneous pre-finetuning on linguistically relevant tasks, such as finetuning on CA texts, POS tagging, segmentation, and text diacritization, all framed as MLM tasks. This enriches the model's contextual understanding by integrating knowledge from these tasks.
Following this, PTCAD progresses into a finetuning phase, where ATD is treated as a token classification task. This phase capitalizes on the contextual groundwork laid in the pre-finetuning phase, refining the model's ability to accurately diacritize Arabic text. Our rigorous evaluation of PTCAD on two widely recognized benchmark datasets — the cleaned version of Tashkeela provided by \citet{multi_comp} and the version by \citet{atd_dnn} — showcases its effectiveness. Remarkably, PTCAD achieves a 20\% reduction in WER compared to the state-of-the-art models on both benchmarks, underlining the success of our methodology.
Furthermore, through an ablation study, we emphasize the pivotal role of the pre-finetuning phase in shaping the overall performance of PTCAD. 

Moreover, our comprehensive error analysis offers crucial insights into the types and sources of errors, guiding future refinements and developments in Arabic diacritization methods. Alongside these primary evaluations, we conducted an assessment of GPT-4's capability in the ATD task. In contrast to PTCAD, GPT-4 exhibited lower performance levels, with a DER of 20\% and a WER of 30\% on the benchmark dataset provided by \citet{atd_dnn}. This comparison underscores the specialized efficiency of PTCAD in handling the complexities of Arabic diacritization compared to general-purpose models like GPT-4.
So, the main contributions of our paper can be succinctly summarized as follows:

\begin{itemize}
\item Introduction of PTCAD, a two-phase training methodology for the ATD task. The first phase, pre-finetuning, integrates learning from linguistically relevant tasks to enhance contextual understanding. The second phase involves finetuning ATD as a token classification task for pre-trained BERT-like models.

\item Effectiveness of PTCAD demonstrated through benchmarking on two standard datasets: the cleaned versions of Tashkeela by \citet{multi_comp} and \citet{atd_dnn}. Notably, PTCAD achieves a 20\% reduction in WER compared to the state-of-the-art (SOTA) on both datasets. Additionally, in our assessments, GPT-4 showed comparatively lower performance in the ATD task.

\item An ablation study revealing the efficacy of the pre-finetuning phase, highlighting the importance of multi-task learning in Phase 1 for overall performance enhancement. The study is complemented by an error analysis to identify and address sources of errors, further substantiating our training strategy's effectiveness.
\end{itemize}

The rest of the paper is organized as follows. Section 2 provides a comprehensive review of relevant literature on Arabic diacritization. Section 3 introduces our proposed modeling approach. Section 4 presents the datasets used and the evaluation metrics employed in our study. In Section 5, we present the experimental results, including an ablation study that assesses the significance of multi-task training. Section 6 presents the advantages and limitations of our approach. Finally, the last section summarizes the study's key findings and discusses future research directions.

\section{Related Work}\label{Related Work}
The development of automatic diacritization of Arabic text has been explored in various studies. This survey categorizes related works into two groups based on the type of Arabic language data used, namely CA and MSA. Within each category, the methods are arranged chronologically and benchmarked on similar datasets to present a clear progression of the evolving research in this domain.
\\

\noindent
\textbf{Focusing on CA (Table \ref{surveyCA}).} \citet{multi_comp}  utilized the cleaned version of the Tashkeela dataset \citep{multi_comp}, achieving a DER of 3.39\% and a WER of 9.94\%. This result was reached using a deep learning model composed of a multi-layer RNN with LSTM and dense layers, combined with rule-based and statistical correction components.
\citet{effective_dl_ad} reported enhanced results on the same dataset, obtaining a WER of 4.43\% and DER of 1.13\%. They employed a character-level Convolutional Bank and Highway network architecture followed by a Bidirectional GRU module. Subsequently, \citet{simple_ext} used a subset of the data from \citet{multi_comp} and implemented a model with character-level embeddings of size 128 and 4 Bi-LSTM hidden layers. They divided diacritics into four groups and applied a sliding window operation for data generation, resulting in a DER of 3-3.6\% and a WER of 8.55-8.99\%, considering both the inclusion and exclusion of diacritics. \citet{classif_diac} and \citet{transfer_ad_poet}  used the Arabic Poem Comprehensive Dataset 2 (APCD2) \citep{apcd2} for training. The former study employed the model developed by Abandah and Abdel-Karim \citep{abanda}, achieving significant results with a WER of 20.40\% and a DER of 6.08\% on a specific test subset of APCD2. This subset consisted of text samples with a diacritic-to-letters ratio of 50\% or higher. On the other hand, the latter study augmented diacritization models with a meter classification model.
For the first test subset with a diacritics-to-letters ratio of 50\% or higher, they achieved a DER of 4.46\% and a WER of 15.43\%. Furthermore, they specifically focused on a separate test subset of APCD2 containing text samples with a diacritics-to-letters ratio of 67\% or higher, obtaining a DER of 3.54\% and a WER of 12.34\%.

Moving on to additional datasets, \citet{atd_dnn} benchmarked on a cleaned version of Tashkeela with 55k sentences and 2.3 M words. They achieved a DER of 4.36\% and WER of 10.89\% using the Shakkala model \citep{shakkala} which is a character-level Bi-LSTM model. \citet{fadel-etal-2019-neural} also benchmarked on the \citep{atd_dnn} dataset, achieving a DER of 3\% and a WER of 7.39\%. They employed a character-level RNN with a Bidirectional Neural Grammar (BNG) module. The architecture included 2 BiCuDNNLSTM layers with 512 hidden units each, and the model was trained for 10 epochs. \citet{efficient_hier} benchmarked on the same data and achieved a DER of 2.09\% and WER of 5.08\% using both character and word levels. Their architecture involved two levels: words and characters. The model first understood the context of the whole sentence by looking at the sequence of words. Then, for each word, it broke down the word into its characters and paid attention to the relationship between each character and the context of all the words in the sentence.
Using this same data, \citet{lamad} achieved a DER of 2.71\% and WER of 6.9\% with an architecture incorporating a novel linguistic feature representation, a Bi-LSTM layer to learn character-level linguistic features, and an attention mechanism to extract the most effective linguistic features. Their work also benchmarked on the Holy Quran (6k sentences), and Sahih Al-Bukhary (9k sentences) and achieved a DER of 2.7\% and WER of 6.56 \% for the Holy Quran and a DER of 2.52\% and WER of 5.20\% for Sahih Al-Bukhary.
\\

\noindent
\textbf{Focusing on MSA (Table \ref{surveyMSA}).} \citet{joint_diac_lemma} adopted a joint modeling approach employing a sequence-to-sequence architecture with distinct parameter-sharing strategies for MSA and Egyptian dialects. Their work was benchmarked on PATB (parts 1, 2, and 3) \citep{patb} for MSA and the ARZ dataset (parts 1-5) \citep{arz} for the Egyptian dialect. The study achieved an impressive accuracy of 93.9\% in diacritized forms. \citet{multi_task_ad} proposed a multi-task learning model to jointly optimize diacritic restoration along with related NLP tasks like word segmentation, POS tagging, and syntactic diacritization. The paper's model was benchmarked on PATB (parts 1, 2, and 3) following the same data division as \citep{diab2013}. It yielded promising results, achieving a WER of 7.51\% and a DER of 2.54\%. \citet{recent_ad} introduced a combination of LSTM networks and Maximum Entropy methods, employing knowledge distillation techniques. The study was benchmarked on PATB part 3 \citep{patb}, with training conducted on 600 stories and testing on 91 articles from Al Nahar News text \citep{annahar}. The model achieved an impressive WER of 4.3\%. Subsequently, \citet{advers_ad} explored the use of an adversarial training strategy in the conventional sequence-to-sequence model. A DER of 2.15\% and a WER of 6.35\% were achieved on the PATB dataset.  \citet{diac_trans} proposed a multitask learning approach employing a character-level transformer encoder-decoder model for diacritization and parallel text translation, benchmarked on PATB parts 1, 2, and 3, \citep{patb} resulting in a WER of 4.79\%. \citet{highly_eff} and \citet{system_diac} utilized a compilation of 4.5 million words from \citep{darwish-etal-2017} as an MSA train set and a corpus extracted from the WikiNews dataset, consisting of 18.3 thousand words, as an MSA test set.
\citet{highly_eff} employed a character-level seq2seq model on a sliding window of words, achieving a WER of 4.49\% and a DER of 1.21\%. In contrast, \citet{system_diac} used a similar approach but included a voting component to select the most common diacritized form for each word, yielding a WER of 4.5\%. Finally, in \citep{ad_feature_rich}, a feature-rich, sequence-to-sequence model was used to restore diacritics in Arabic text. The model achieved a DER of 0.9\% and a WER of 2.9\% by training on the corpus used to train the RDI \citep{RDI} diacritizer and the Farasa diacritizer \citep{darwish-etal-2017}, and evaluating it on the WikiNews dataset.
\\
\\
The comprehensive summaries of all the relevant elements mentioned in the survey for  CA and MSA can be found in Table \ref{surveyCA} and Table \ref{surveyMSA} respectively.

\begin{landscape}

\begin{table}[p]
\centering
\resizebox{560pt}{!}{%
\centering
\begin{tblr}{
  row{1} = {c},
  cell{7}{3} = {r=4}{},
  hline{2,5,7} = {-}{},
  hline{3-4,6} = {2-5}{},
  hline{8-10} = {2,4-5}{},
  hline{11} = {1-4}{},
}
\centering
\textbf{Benchmark Data}                                           & \textbf{Article}               & \textbf{Details on data}                                                                                                                                                                                                                             & \textbf{Experimental details and results}                                                                                                      & \textbf{Approach}                                                                                                                                                                                                     \\
                                                               & \citet{multi_comp}             & {Train: 28M words\\ Test: 1.7M words}                                                                                                                                                                                                                & {DER of 3.39\%\\ WER of   9.94\%\\  Including partially diacritized sentences}                                                                 & {A model composed of a multi-layer RNN with LSTM and Dense layers,\\ combined with rule-based and statistical correction components.}                                                                                 \\
{\citet{multi_comp}\\ cleaned version of\\ Tashkeela}          & \citet{effective_dl_ad}        & {Additional   post-processings\\ Train: 2.3M sentences\\ Test: 124K sentences}                                                                                                                                                                       & {DER of 1.13 \%\\ WER of 4.43\%\\  Excluding partially diacritized sentences}                                                                  & {Character-level Convolutional Bank and Highway Network architecture\\ followed by a Bidirectional GRU module.}                                                                                                       \\
                                                               & \citet{simple_ext}             & {A subset of \citep{multi_comp} dataset\\ Train: the first 10 files from the train set\\ Test: the first file from both validation\\ and test set}                                                                                                   & {DER of 3.6\%\\ WER of 8.55\%\\  Excluding partially diacritized sentence}                                                                     & {Model - employing 128-sized character-level embeddings and four Bi-LSTM \\ hidden layers - categorizes diacritics into four groups utilizing a sliding\\ window for data generation.}                                \\
{Arabic Poem\\ Comprehensive Dataset 2\\(APCD2) \citep{apcd2}} & \citet{transfer_ad_poet}       & {Cleaning APCD2:\\ Removing partially diacritized sentences\\ up to a certain   threshold\\  DS1: (Train: 313K verses, Test: 55K verses)\\  DS2: (Train: 76K verses, Test: 13K verses)}                                                              & {DS1 test: DER of 4.46\%\\ WER of 15.43\%\\ DS2 test: DER of 3.54 \%\\ WER of 12.34\%}                                                         & {Enhanced diacritization models combined with a meter classification model\\trained first on DS1 (Diacritics to letters ratio \textgreater{}= 50\%) then on DS2 (Diacritics to\\letters ratio \textgreater{}= 67\%).} \\
                                                               & \citet{classif_diac}           & {Selecting from APCD2 all the verses in the\\ training set that have diacritics to letters \\ ratio of 0.50 or higher\\ 368K diacritized verses consisting of 3.5M\\ words which were then split into 85\% training \\ set and 15\% validation set.} & {DER of 6.08\%\\ WER of 20.40\%}                                                                                                               & The model of \citet{abanda}                                                                                                                                                                                           \\
                                                               & \citet{atd_dnn}                & {Train: 2M words and 50K sentences\\ Test: 107K words and 2.5K sentences}                                                                                                                                                                            & {DER of 4.36\%\\ WER of  10.89\%\\  Excluding partially diacritized sentences}                                                                 & {Shakkala model employs Bi-LSTM networks and character embeddings,\\ iteratively trained on the Tashkeela corpus, discarding detrimental data.}                                                                       \\
                                                               & \citet{fadel-etal-2019-neural} &                                                                                                                                                                                                                                                      & {DER of 3\%\\ WER of 7.39\%\\  Excluding partially diacritized sentences}                                                                      & {Character-level RNN with BNG module, consisting of 2 BiCuDNNLSTMs \\ layers with 512 hidden units each.}                                                                                                             \\
{\citet{atd_dnn},\\ Data}                                      & \citet{efficient_hier}         &                                                                                                                                                                                                                                                      & {DER of 2.09\%\\ WER of 5.08\%\\  Excluding partially diacritized sentences}                                                                   & {Two-level architecture analyzes sentence context via word sequences,\\ then dissects each word into characters, examining the relationship\\ between each character and the sentence's word context.}                \\
                                                               & \citet{lamad}                  &                                                                                                                                                                                                                                                      & {DER of 2.71\%\\ WER of 6.9\%\\  Excluding partially diacritized sentences}                                                                    & {Architecture integrates novel linguistic feature representation, a Bi-LSTM\\ layer for character-level linguistic feature learning, and an attention mechanism\\ to extract prominent linguistic features.}          \\
{The Holy Quran\\ and Sahih Al-Bukhary}                        & \citet{lamad}                  & {The Holy Quran: 6K   sentences\\  Sahih Al Bukhary: 9K sentences}                                                                                                                                                                                   & {The Holy Quran: DER of 2.7\%\\ WER of 6.56 \%\\ Sahih Al-Bukhary: DER of 2.52\%\\ WER of 5.20\%\\  Excluding partially diacritized sentences} & Idem.                                                                                                                                                                                                                 
\end{tblr}
}
\caption{Benchmarking Tashkeel literature on CA}
\label{surveyCA}
\end{table}

\end{landscape}

\begin{landscape}
\begin{table}[]
\centering
\resizebox{570pt}{!}{%
\begin{tabular}{cllll}
\textbf{Benchmark Data} &
  \multicolumn{1}{c}{\textbf{article}} &
  \multicolumn{1}{c}{\textbf{Details on Data}} &
  \multicolumn{1}{c}{\textbf{Experimental details and results}} &
  \multicolumn{1}{c}{\textbf{Approach}} \\ \hline
   &
  \begin{tabular}[c]{@{}l@{}}\citet{joint_diac_lemma}\end{tabular} &
  \begin{tabular}[c]{@{}l@{}}PATB parts 1,2, and 3 \citep{patb}\\ follow the same data division as\\  \citet{diab2013} \\ Train : 502K Words\\ Test : 64K Words\end{tabular} &
  \begin{tabular}[c]{@{}l@{}}Diacritized forms accuracy\\ of 93.9\% \\ (The accuracy of the diacritized \\ form of the words).\end{tabular} &
  \begin{tabular}[c]{@{}l@{}}Sequence-to-sequence  architecture with diverse parameter sharing strategies.\\ Lexicalized features (lemmas, diacritized forms) are modeled at the character-level,\\ while   non-lexicalized features (gender, number) are modeled at the word-level.\\ The model employs multitask-learning for non-lexicalized features\\ and separate decoders for lexicalized features, aiming to enhance context modeling\\  and disambiguation of ambiguous lexical choices.\end{tabular} \\ \cline{2-5} 
 &
    \citet{multi_task_ad} &
   &
  \begin{tabular}[c]{@{}l@{}}DER of 2.54\%\\ WER of 7.51\%\end{tabular} &
  \begin{tabular}[c]{@{}l@{}}Multi-task learning model to jointly optimize diacritic restoration along with\\ related NLP tasks like word   segmentation, POS tagging,\\ and syntactic diacritization\end{tabular}\\ \cline{2-2} \cline{4-5} 
 \multirow{-5}{*}{\begin{tabular}[c]{@{}c@{}}PATB \citep{patb}\end{tabular}} &
  \citet{recent_ad} &
  \begin{tabular}[c]{@{}l@{}}PATB Part 3, version2\\ train: 600 stories (340,281 words)\\ from the Al Nahar News text \citep{annahar}.\\ test: 91 articles (about 52,000 words)\\ from October to December 2002\end{tabular} &
  WER of 4.3\% &
  \begin{tabular}[c]{@{}l@{}}Combined LSTM networks and Maximum Entropy methods\\ with knowledge distillation techniques\end{tabular} \\ \cline{2-5} 
 &
  \citet{advers_ad} &
   \multirow{-2}{*}{\begin{tabular}[c]{@{}l@{}}PATB parts 1,2, and 3\\ follow the same data division as\\ \citet{diab2013}\\ Train : 502K Words\\ Test : 64K Words\end{tabular}} &
  \begin{tabular}[c]{@{}l@{}}DER of 1.77\%\\ WER of  4.88\%\end{tabular} &
  \begin{tabular}[c]{@{}l@{}}Combination of regularized decoding and adversarial training.\\ In addition to the gold diacritized   sentences, the model synthetize new sentences,\\ and train on both gold and   synthetic ones and the descriminator tries to predict\\ whether a sentence is a   gold one or not\end{tabular}\\ \cline{2-2} \cline{4-5} 
 &
  \begin{tabular}[c]{@{}l@{}}\citet{diac_trans}\end{tabular} &
   &
  WER of  4.79\% &
  \begin{tabular}[c]{@{}l@{}}Multitask learning approach employing a character-level transformer\\ encoder-decoder model for diacritization and parallel text translation\end{tabular} \\ \hline
\multicolumn{1}{l}{\multirow{-2}{*}{\begin{tabular}[c]{@{}l@{}}MSA train\\ \citet{darwish-etal-2017}\\ MSA test: WikiNews corpus \citep{darwish-etal-2017} \end{tabular}}} &
  \begin{tabular}[c]{@{}l@{}}\citet{highly_eff}\end{tabular} &
   \multirow{-2}{*}{\begin{tabular}[c]{@{}l@{}}Train : 4.5 M words\\ Test : 18K   words\end{tabular}} &
  \begin{tabular}[c]{@{}l@{}}DER of 1.21\%\\ WER of   4.49\%\\  Including partially\\ diacritized sentences\end{tabular} &
  \begin{tabular}[c]{@{}l@{}}Character level seq2seq model on a sliding window of words that are\\ represented using characters, and we employ voting to pick the best\\ most likely diacritized form from different   windows.\end{tabular} \\ \cline{2-2} \cline{4-5} 
 &
  \begin{tabular}[c]{@{}l@{}}\citet{system_diac}\end{tabular} &
   &
  \begin{tabular}[c]{@{}l@{}}WER of   4.5\%\\ \\  Including partially\\ diacritized sentences\end{tabular} &
  \begin{tabular}[c]{@{}l@{}}Character-level sequence-to-sequence model. After diacritization, the system \\ includes a voting component to select the most common diacritized form\\ for each word, considering multiple diacritized versions obtained from\\  consecutive windows of the text.\end{tabular} \\ \hline
\multicolumn{1}{l}{\begin{tabular}[c]{@{}l@{}}MSA train: the corpus used\\ to train the RDI diacritizer \citep{RDI} \\ and the Farasa diacritizer \citep{darwish-etal-2017} \\  MSA test: WikiNews corpus \citep{darwish-etal-2017} \end{tabular}} &
  \begin{tabular}[c]{@{}l@{}}\citet{ad_feature_rich}\end{tabular} &
  \begin{tabular}[c]{@{}l@{}}9.7M tokens with approximately\\ 194K unique surface forms\\ (excluding numbers \\ and punctuation marks)\end{tabular} &
  \begin{tabular}[c]{@{}l@{}}DER of 0.9\%\\ WER of   2.9\%\\  Including partially\\ diacritized sentences\end{tabular} &
  \begin{tabular}[c]{@{}l@{}}Two separate Deep Neural Network architectures to recover both kinds\\ of diacritic types, core-word (CW) diacritics, and case endings (CEs).\\ For CW diacritics, they used a character-level biLSTM model with\\ associated features, informed using word segmentation information\\ and a unigram language model as post corrector. For CE recovery,\\ they employed a word-level biLSTM model that is trained with a rich set\\ of surface, morphological, and syntactic features.\end{tabular}
\end{tabular}
}
\caption{Benchmarking Tashkeel literature on MSA}
\label{surveyMSA}
\end{table}
\end{landscape}

\section{Models \& The Proposed Approach}
In this section, we present PTCAD, our novel approach to ATD. We will begin by presenting the two-phase training approach, starting with a detailed examination of the initial pre-finetuning phase on tasks closely related to ATD. Subsequently, we will delve into the second phase, framing ATD as a token classification finetuning task. We will outline the specifics of this process and discuss its impact on model performance

\subsection{PTCAD Overview}\label{overv}

PTCAD can be broadly summarized in the following two phases:

\begin{enumerate}
    \item  \textbf{Phase 1: Pre-finetuning on Linguistically Relevant Tasks.} Recognizing the intertwined nature of various linguistic tasks, our method incorporates a pre-processing phase that targets linguistically pertinent tasks. Before diving into ATD-specific training, the model undergoes simultaneous pre-finetuning on a suite of relevant tasks. These encompass multi-task finetuning on CA texts, POS tagging, segmentation, and ATD itself. Remarkably, all these tasks are formulated as MLM undertakings, streamlining the training process. This holistic training regimen ensures that the model is equipped with a rich contextual understanding, a facet crucial for diacritization.
    
    \item \textbf{Phase 2: Reframing ATD as a Token Classification Task.} In the finetuning phase, we pivot from traditional methods and envision ATD as a finetuning task for BERT-like models. By casting ATD in this new light, we tap into the model's capacity to effectively comprehend and integrate contextual cues, leading to marked advancements in diacritization accuracy. To align with this vision, ATD is treated analogously to token classification tasks, reminiscent of Named Entity Recognition (NER). This involves a pivotal pre-processing step, converting the native input sentence structure to a format amenable to token classification.

\end{enumerate}

\begin{figure}[H]
    \centering
    \includegraphics[width=0.8\textwidth]{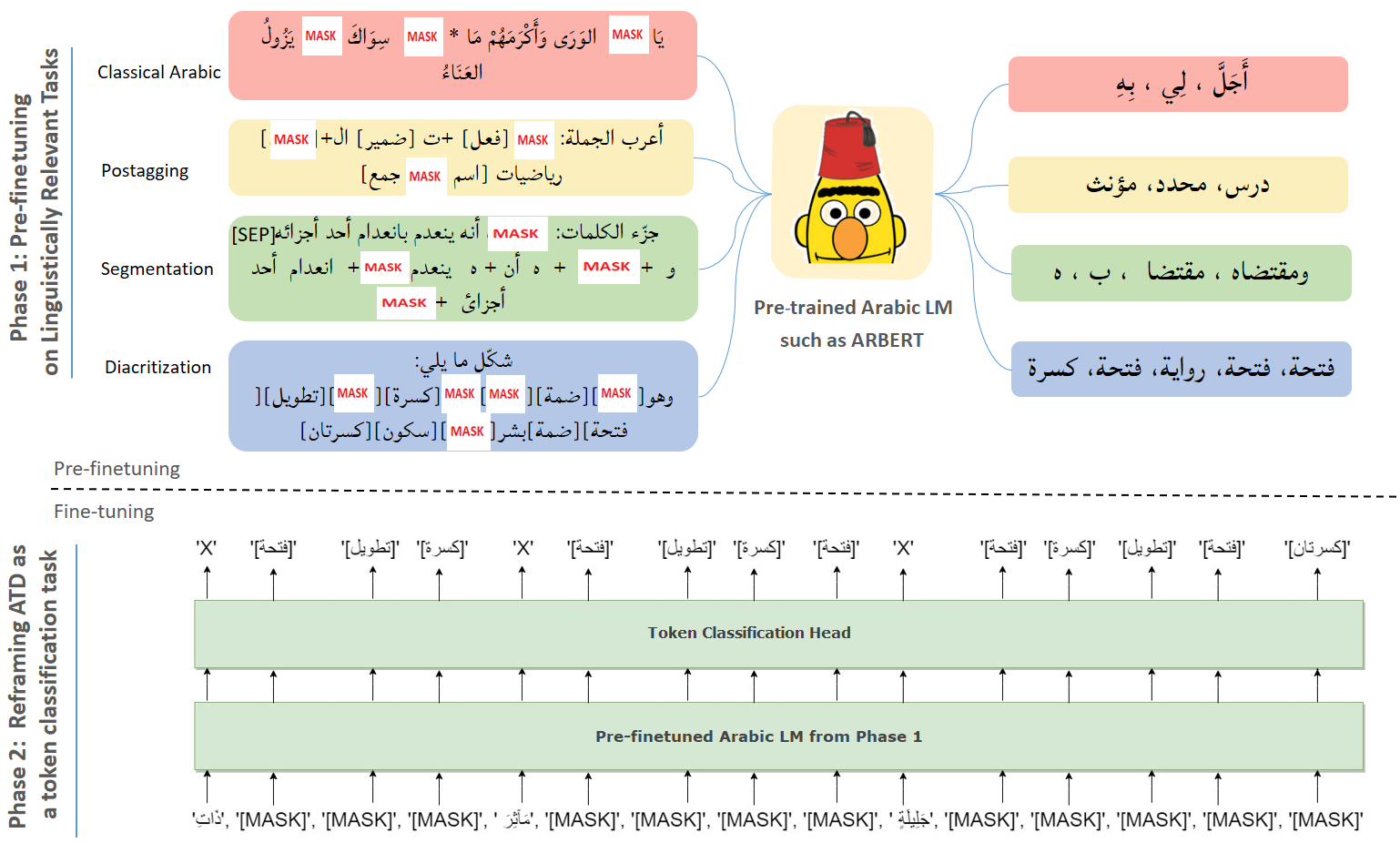}
    \caption{Global overview of PTCAD approach}
    \label{overview}
\end{figure}

Figure \ref{overview} illustrates our approach.

The versatility of the PTCAD framework allows for diverse implementations, accommodating different linguistic task combinations. These variations form the basis of the ablation study we will be presenting later on, with the hypothesis that the FULL model (PTCAD) will demonstrate superior performance, underscoring the importance of a comprehensive pre-finetuning phase in the PTCAD framework. Possible variations of PTCAD can be:
\begin{itemize}
    \item \textbf{FULL Model (PTCAD):} Two-phase training approach with the pre-finetuning phase incorporating all four pre-finetuning tasks (CA texts, POS tagging, segmentation, and ATD). This complete model serves as the benchmark for optimal performance in our study.
    \item \textbf{CA-POS-Seg Model:} Two-phase training approach with the pre-finetuning phase including three pre-finetuning tasks (CA, POS tagging, and segmentation), omitting the ATD task. It assesses the impact of excluding direct ATD pre-finetuning.
    \item \textbf{CA-POS Model:} Two-phase training approach with the pre-finetuning phase focusing on two tasks (CA and POS tagging), removing segmentation in addition to ATD. This model tests the core linguistic capabilities without the complexity of segmentation.
    \item \textbf{CA Model:} Two-phase training approach with the sole CA pre-finetuning task, providing insight into the foundational impact of CA texts on ATD performance.
    \item \textbf{TCO Model (Token Classification Only Model):} Single-phase training approach excluding all pre-finetuning tasks, relying solely on the token classification finetuning phase. This model serves to highlight the importance of the holistic pre-finetuning phase.
\end{itemize}
Through these variations, we aim to systematically dissect the contributions of individual tasks within the PTCAD framework, thereby showcasing the efficacy and necessity of each related task in the full model.

\subsection{Phase 1: Pre-finetuning on Linguistically Relevant Tasks}

Our strategy involves an initial pre-finetuning of the BERT-like model on tasks directly relevant to diacritization. Specifically, we simultaneously finetune the model on CA text then POS tagging, Text Segmentation, and finally on Automatic Diacritization. Each of these is framed as an MLM task to facilitate training. This pre-finetuning allows the model to more effectively interpret and incorporate contextual cues, which in turn substantially improves its performance in ATD.
\begin{itemize}
    \item The CA pre-finetuning task is straightforward. We feed the model with plenty of CA texts to be trained on using the MLM objective.

\setcode{utf8}
$$\RL{إذ نازلتنا الليالي فيه عن كثب}$$

    \item For POS tagging, we utilize an instruction-based approach, framing each POS tagging sample as an instruction. The modified input is then fed into the model for finetuning the Masked Language Model. Here's a sample instruction:

\setcode{utf8}
\begin{eqnarray*}
    & \RL{أعرب الجملة: إذ [حرف] نازلتنا [اسم مذكر فردي]}\\
    & \RL{ال+ [محدد] ليالي [اسم مؤنث جمع] في [ظرف جر]}\\
    & \RL{ +ه [ضمير] عن [ظرف جر] كثب [اسم مذكر فردي]  }\\
\end{eqnarray*}
The meaning of the added task prefix $\RL{أعرب الجملة}$ is "Classify the words in the following sentence". This is concatenated with the input sentence where we write the POS after each word.

    \item For text segmentation, a similar instruction-based approach is employed. A segmentation task sample is presented as an instruction, and the MLM is finetuned accordingly. Here's an instance of such instruction:
\begin{eqnarray*}
    & {[}SEP{]}\RL{ أحد أجزائه } \RL{ جزّء الكلمات: ومقتضاه أنه ينعدم بانعدام}\\
    & \RL{و+مقتضا+ه أن+ه ينعدم ب+انعدام أحد أجزائ+ه }
\end{eqnarray*}
The meaning of the added task prefix $\RL{جزّء الكلمات}$ is "Segment the following words." The input sentence is then concatenated with "Segment the words" followed by the [SEP] token. The output sentence contains the segmented words, with each segment separated by the "+" symbol.

    \item The same goes for text diacritization pre-finetuning task. Here's an instance of such instruction: 
\begin{eqnarray*}
    & \RL{شكّل ما يلي: وهو[فتحة][ضمة][فتحة]}\\
    & \RL{رِواية[كسرة][فتحة][تطويل][فتحة][ضمة]}\\
    & \RL{بِشرٍ[كسرة][سكون][كسرتان]}
\end{eqnarray*}
The meaning of the added task prefix $\RL{شكّل ما يلي}$ is "diacritize the following". This is concatenated with the input sentence where we write the sequence of diacritization marks after each word.
\end{itemize}
By this dual-phase approach, we leverage the strength of BERT-like models in contextual understanding, simultaneously enriching their performance with linguistic insights related to diacritization. This fusion results in a model that shows improved performance in ATD, with a fine-grained understanding of language nuances and a highly accurate diacritic assignment mechanism.
\subsection{Phase 2: Reframing ATD as a Token Classification Task}

In this subsection, we introduce our innovative method for addressing ATD by adapting it into a finetuning process for existing BERT-like models. 

\begin{figure}[H]
    \centering
    \includegraphics[width=0.9\textwidth]{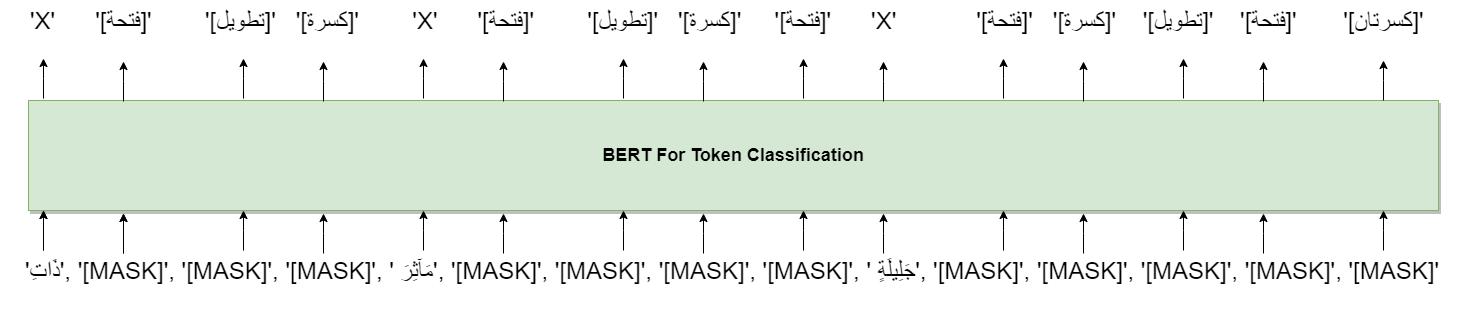}
    \caption{Sample input transformation for ATD}
    \label{tashkeeltokenclassif}
\end{figure}

Key to this method is an essential pre-processing stage, which transforms the original sentence structure into a format suitable for token classification.
\\
\\
\noindent
\textbf{Input Transformation for ATD.} Our methodology treats ATD as a token classification task, a process similar to  Named Entity Recognition (NER). The initial step involves transforming the raw input sentence into a form suitable for token classification. Each sentence to be diacritized is processed by inserting a number of [MASK] tokens after each word, equivalent to the number of Arabic letters contained in the word. This arrangement allows each mask token to correspond to a distinct Arabic letter in the preceding word. Each mask token then undergoes token classification through the BERT model, producing softmax probabilities for each diacritical mark. This is represented in a 'labels' sequence mirroring the 'tokens' sequence. Figure \ref{tashkeeltokenclassif} illustrates an example of such a sample for the Token Classification Task.

Note that the Transformers library \citep{hf_transformers} uses the label ID -100 to indicate that the loss associated with a label should not be taken into account. This is the case for the label "X".

\section{Experimental Setup}
In this section, we detail our methodology for evaluating PTCAD, starting with the dataset used for pre-finetuning our model. Subsequently, we describe the finetuning dataset, outline the hyperparameters chosen for our experiments, and specify the evaluation metrics used to assess the model's performance.

\subsection{Pre-finetuning Datasets}
The datasets used in this phase come from two primary sources: the OpenITI-proc corpus \citep{openiti} and the cleaned version of Tashkeela by \citet{multi_comp}. Table \ref{prefinetuning_data} provides a detailed breakdown of the specific subsets used for each task.
\\
\\
\textbf{OpenITI corpus.} This corpus is part of the broader Open Islamicate Texts Initiative (OpenITI), which focuses on compiling texts in Arabic, Persian, and other languages, largely from religious and literary traditions across various historical periods. The main sources of these texts are Al-Maktaba Al-Shamela (commonly known as Shamela), the Shia online library \citep{shialibrary}, and Al-Jami’ Al-Kabir (JK). The OpenITI-proc corpus, a processed version of the original OpenITI, encompasses several specialized datasets, including but not limited to:

\begin{itemize}
\item The OpenITI.sentences dataset: A subset from the main OpenITI dataset, specifically processed for extracting sentences.
\item The OpenITI.pos dataset: A subset of the main OpenITI dataset, processed to provide Part-of-Speech (POS) tagging information.
\item The OpenITI.segmentation dataset: A subset derived from the main OpenITI dataset, processed for text segmentation purposes.
\end{itemize}
These datasets are essential for various tasks such as providing raw text, POS tags, and sentence segmentation.
\\
\\
\textbf{Pre-processing and Annotation.}
In its preprocessing stage, OpenITI utilizes a combination of OpenNLP and the Farasa toolkit. The key tasks automated using the Farasa toolkit include:

\begin{itemize}
    \item \textbf{Morphological Segmentation}: Decomposing words into their constituent morphemes using SVMrank with a linear kernel.
    \item \textbf{POS tagging}: Applying the simplified PATB tagset for determining the appropriate tag for each morpheme, including gender and number aspects.
\end{itemize}
While sentence splitting was performed using OpenNLP, trained on specific Arabic corpora, Farasa was predominantly used for other preprocessing activities. The datasets utilized in the pre-finetuning stage of PTCAD are concisely summarized in Table \ref{prefinetuning_data}.

\begin{table}[H]
\centering
\resizebox{\textwidth}{!}{%
\begin{tabular}{l|l|c}
\hline
\textbf{Pre-finetuning task} & Data name                                          & \multicolumn{1}{l}{\textbf{Data size}}                                                                                    \\ \hline
CA                  & OpenITI.sentences                       & \multirow{4}{*}{\begin{tabular}[c]{@{}c@{}}1 million sentences randomly chosen\\ from each dataset\end{tabular}} \\ \cline{1-2}
POS tagging         & OpenITI.pos                             &                                                                                                                  \\ \cline{1-2}
Segmentation        & OpenITI.segmentation                   &                                                                                                                  \\ \cline{1-2}
Diacritization      & \citet{multi_comp} cleaned version of Tashkeela &                                                                                                                  \\ \hline
\end{tabular}
}
\caption{Datasets used for the Pre-finetuning phase}
\label{prefinetuning_data}
\end{table}

\subsection{Finetuning Datasets}
To assess PTCAD's performance in CA, this study utilizes two extensively cleaned versions of the Tashkeela corpus, hereafter referred to as  \textbf{"Abbad Tashkeela"} and \textbf{"Fadel Tashkeela"} for clarity throughout the rest of paper. The dataset named "Abbad Tashkeela," provided by \citet{multi_comp}, and the one denoted as "Fadel Tashkeela," provided by \citet{atd_dnn}, both offer high-quality, diacritized corpora extensively used in previous research. Table \ref{finetuning_data} outlines the key statistics of these datasets.

\begin{table}[H]
\centering
\resizebox{\textwidth}{!}{%
\begin{tabular}{l|l}
\hline
Data                                            & Data size                                                                                                        \\ \hline
\citet{multi_comp} cleaned verison of Tashkeela & \begin{tabular}[c]{@{}l@{}}Train: 28M words\\ Test: 1.7M words\end{tabular}                                      \\ \hline
\citet{atd_dnn} cleaned version of Tashkeela    & \begin{tabular}[c]{@{}l@{}}Train: 2M words and 50K sentences\\ Test: 107K words and 2.5K sentences\end{tabular} \\ \hline
\end{tabular}
}
\caption{Datasets used for the finetuning phase}
\label{finetuning_data}
\end{table}

\noindent
\textbf{Preprocessing.} From the test sets of these two datasets, we excluded the partially diacritized sentences having a ratio of diacritized words to a total number of words less than 90\%. A word is diacritized if all its letters are diacritized except these letters or a combination of letters:
$$\RL{ى، ئ، و، ا، ي، إ، ؤ، ال}$$
$$\RL{حرف + و، حرف + ا، حرف + ى، حرف + ي}$$
Following the constraints of the base model (BERT), which supports a maximum sequence length of 512 tokens, sentences in the \textit{test set} exceeding this length were recursively segmented using line breaks, periods, commas, and ultimately spaces, ensuring that each resulting subsentence contained fewer than 512 tokens. An important distinction in processing the \textit{training set} should be highlighted: segments created by splitting sentences that exceeded 512 tokens were not utilized in training the model.

\subsection{Metrics}
To assess the effectiveness of the ATD system, two primary evaluation metrics were employed: DER and WER.
\begin{itemize}
    \item DER measures the proportion of characters that have been incorrectly labeled with diacritics. It indicates the accuracy of the diacritization process and the system's ability to predict the correct diacritic for each character.
    \item WER calculates the percentage of words in which at least one letter has been incorrectly diacritized. This metric allows for a more comprehensive evaluation of the system's performance at the word level, considering the impact of diacritization errors on overall text understanding.
\end{itemize}

\subsection{Hyperparameters}

The adopted training configuration adheres primarily to the default settings as outlined in the "Transformers Training Arguments" \citep{training_args}. However, we've made specific adjustments to better suit our requirements. These customizations include setting the batch size to 64 and the maximum sequence length to 512. Additionally, the number of training epochs varies depending on the training phase and the specific benchmark being utilized. The table below succinctly presents the number of epochs assigned for each phase and benchmark:

\begin{table}[H]
\centering
\resizebox{\textwidth}{!}{%
\begin{tabular}{c|c|ll}
\hline
\multirow{2}{*}{}               & \multirow{2}{*}{Pre-finetuning phase} & \multicolumn{2}{c}{finetuning phase}                 \\ \cline{3-4} 
                                &                                       & \multicolumn{1}{l|}{Abbad Tashkeela} & Fadel Tashkeela \\ \hline
\multicolumn{1}{l|}{Nb epochs} & \multicolumn{1}{l|}{20}               & \multicolumn{1}{l|}{10}              & 40 \\ \hline            
\end{tabular}
}
\caption{Number of epochs per training phase}
\label{nb_epo}
\end{table}

\section{Experimental results}\label{exp}
In this section, we present a detailed exploration of the experimental results derived from our PTCAD methodology. Our experiments are carefully structured to assess PTCAD's performance across various dimensions.

\subsection{Finding the best Pre-trained Language Model}
In this section, we evaluate several Arabic MLM-style pre-trained models for their suitability in ATD. Key models, such as AraBERT (in various versions, AraBERTv1, AraBERTv02), MARBERT \citep{arbert}, Arabert \citep{arabert}, and Camelbert \citep{camelbert}, are considered for their proven quality in Arabic language processing. Due to the large number of models being compared and the computational intensity of the task, we have chosen to benchmark these models solely on the Fadel Tashkeela benchmark, which is favored for its reduced size compared to the other benchmarks that are more resource-intensive.

The focus is on determining which model best adapts to ATD nuances after finetuning, thereby identifying the most effective base for our PTCAD system.
The results are reported in Table \ref{which-pretrained}. The experiments showed that the best pre-trained model to adopt is ARBERT.

\begin{table}[H]
\centering
\begin{tabular}{l|ll}
\hline
\multicolumn{1}{c|}{\multirow{2}{*}{\textbf{Model Name in HuggingFace}}}           & \multicolumn{2}{c}{Fadel Tashkeela} \\ \cline{2-3} 
\multicolumn{1}{c|}{}                                                              & \textbf{DER}      & \textbf{WER}    \\ \hline
UBC-NLP/ARBERTv2                                                                   & 2.03              & 6.17            \\ \hline
\textbf{UBC-NLP/ARBERT}                                                            & \textbf{2.03}     & \textbf{6.1}    \\ \hline
UBC-NLP/MARBERT                                                                    & 2.4               & 7.48            \\ \hline
UBC-NLP/MARBERTv2                                                                  & 2.13              & 6.56            \\ \hline
\begin{tabular}[c]{@{}l@{}}CAMeL-Lab/bert-base-arabic\\ -camelbert-ca\end{tabular} & 54                & 97.3            \\ \hline
aubmindlab/bert-large-arabertv2                                                    & 55.35             & 94.93   \\\hline       
\end{tabular}

\caption{ATD as Token Classification finetuning for different Arabic Pre-trained Language Models}
\label{which-pretrained}
\end{table}

\subsection{PTCAD Performance}
\label{benchmarking_subsec}

Using the Abbad Tashkeela and Fadel Tashkeela benchmarks, our goal is to assess PTCAD's capabilities in comparison to others, including general-purpose models like GPT-4 and specialized Arabic diacritization models. It is important to note that the performance metrics for the comparative models are directly sourced from existing literature and have not been independently reproduced in our study.

However, as PTCAD utilizes the ARBERT model, \textbf{handling long sentences} poses a challenge due to the inherent limitations of this pre-trained model, particularly its maximum token sequence length of 512 tokens, during the inference phase. This limitation is notably problematic for ATD, as demonstrated in the Fadel Tashkeela benchmark, where over 15\% of sentences exceeded this limit. When such sentences are divided into sub-sentences of fewer than 512 tokens, it is observed that 30\% of the test set comprises these truncated sentences. Several methods have been proposed for managing long sequences in transformer-based models like BERT. These include Sentence Chunking, Contextual Embeddings Concatenation, Attention Masking, Sliding Window, and Hierarchical Models \cite{Longformer, reimers2019sentencebert, Ding2020CogLTXAB, zaheer2021big}, among others. For PTCAD, we explored various inference strategies, particularly employing a slicing window approach for its simplicity.

\begin{itemize}

\item \textbf{PTCAD(0).} In this strategy, we simply split the sentence into segments to diacritize after each 512 tokens, ensuring that each sub-sentence is less than 512 tokens (no overlap). This results in a minimal number of sub-sentences. Subsequently, we diacritize each sub-sentence independently. This approach is computationally efficient but might not fully preserve the contextual flow of the sentences in the inference phase.

\item \textbf{PTCAD(p).}  In this strategy, we take the first window from the beginning of the sentence to nearly 512 tokens, being careful not to split words. After diacritizing this first window, \textbf{we slide it by a constant step of 'p' words} from right to left, forming another sub-sentence of less than 512 tokens. This process continues until we reach the end of the sentence. The advantage of this strategy is the preservation of context. For each word, we look at all the diacritized sentences to which it belongs and take the majority voting of different diacritization forms. This method is particularly effective as it maintains the context by providing different contextual windows for each part of the sentence, which is vital for accurate diacritization.

\end{itemize}
These strategies are designed to mitigate the issues posed by long sentences in Arabic text diacritization, ensuring that our PTCAD model can handle them more effectively without losing contextual meaning, in the inference time.

\begin{table}[H]
\centering
\begin{tabular}{lll}
\hline
\textbf{Model}                         & \textbf{DER}   & \textbf{WER}   \\ \hline
Tashkeela-Model \citep{tashkeelamodel} & 49.96\%         & 96.80\%        \\ \hline
Farasa \citep{farasa}                  & 21.43\%         & 58.88\%         \\ \hline
GPT-4 \citep{gpt4}                     & 20\%            & 30\%            \\ \hline
Mishkal \citep{mishkal}                & 16.09\%         & 39.78\%         \\ \hline
\citet{atd_dnn} Model                  & 4.36\%          & 10.89\%         \\ \hline
Lamad \citep{lamad}                    & 2.71\%          & 6.9\%           \\ \hline
\citet{efficient_hier} Model           & 2.09\%          & 5.08\%          \\ \hline
PTCAD(0) (Ours)                            & \textbf{1.13\%} & \textbf{4.27\%}\\ \hline
PTCAD(20) (Ours)                            & \textbf{1.13\%} & \textbf{4.26\%}\\ \hline
PTCAD(10) (Ours)                            & \textbf{1.12\%} & \textbf{4.24\%}\\ \hline
PTCAD(1) (Ours)                            & \textbf{1.12\%} & \textbf{4.22\%}\\ \hline
PTCAD(5) (Ours)                            & \textbf{1.1\%} & \textbf{4.19\%} \\\hline
\end{tabular}
\caption{Evaluating PTCAD against other ATD models in literature: Benchmarking with the dataset of Fadel Tashkeela. All results are obtained under the conditions of 'with case ending' set to True and 'including diacritics' set to True. The results for the Mishkal and Farasa models are specifically sourced from \citep{atd_dnn}, and the result for the Shakkala model is sourced from \citep{efficient_hier}. Results for all other models are obtained from their respective cited works.}
\label{bench_fadl}
\end{table}

\begin{table}[H]
\centering
\begin{tabular}{lll}
\hline
\textbf{Model}                         & \textbf{DER}   & \textbf{WER}   \\ \hline
Shakkala \citep{shakkala}              & 3.73\%          & 11.19\%        \\ \hline
\citet{simple_ext} Model               & 3.6\%           & 8.55\%          \\ \hline
\citet{multi_comp} Model               & 3.39\%          & 9.94\%          \\ \hline
\citet{effective_dl_ad} Model          & 1.13\%          & 4.43\%          \\ \hline
PTCAD(0) (Ours)                            & \textbf{0.89\%} & \textbf{3.54\%}\\ \hline
PTCAD(20) (Ours)                            & \textbf{0.89\%} & \textbf{3.54\%}\\ \hline
PTCAD(10) (Ours)                            & \textbf{0.89\%} & \textbf{3.54\%}\\ \hline
PTCAD(1) (Ours)                            & \textbf{0.89\%} & \textbf{3.54\%}\\ \hline
PTCAD(5) (Ours)                            & \textbf{0.87\%} & \textbf{3.53\%}\\ \hline
\end{tabular}
\caption{Evaluating PTCAD against other ATD models in literature: Benchmarking with the dataset of Abbad Tashkeela. All results are obtained under the conditions of 'with case ending' set to True and 'including diacritics' set to True. The results for the Mishkal, Farasa, and Shakkala models are specifically sourced from \citep{multi_comp}.}
\label{bench_abbad}
\end{table}

Tables \ref{bench_abbad} and \ref{bench_fadl} summarize the obtained benchmarking results. Our PTCAD model reveals a substantial advancement in ATD. For the Fadel Tashkeela benchmark, PTCAD achieved an impressive 46\% reduction in DER and a 16\% reduction in WER, marking a significant improvement over existing models. In the Abbad Tashkeela benchmark, the model similarly demonstrated an approximate 20\% reduction in both DER and WER. This performance, notably superior to general-purpose models like GPT-4 and other specialized ATD models, is primarily attributed to PTCAD's effective combination of pre-trained BERT-like models and a targeted two-phase training approach. While other models, such as those by \citet{effective_dl_ad} and \citet{efficient_hier}, showed promising results with their unique architectures, PTCAD stands out for its notable error rate reductions. The comparison with GPT-4, which achieved a DER of 20\% and a WER of 30\% in the Fadel Tashkeela benchmark, emphasizes the importance of specialized training in ATD. These findings highlight the diverse methodologies in the field and underscore the potential for continued innovation in ATD solutions.

\subsection{Ablation Study}
In this section, we conduct a series of detailed ablation experiments to meticulously assess the relative importance of each pre-finetuning task in our model's performance. We specifically focus on the PTCAD(0) inference strategy, which involves less computation. The experiments are strategically designed to incrementally incorporate various pre-finetuning tasks, allowing us to isolate and understand the impact of each task on the model's accuracy in ATD.

Our initial baseline involved finetuning solely on the downstream ATD task, without any pre-finetuning, which provided a fundamental performance metric. Subsequently, we introduced pre-finetuning tasks one by one, starting with CA data, followed by adding POS Tagging, and then incorporating text segmentation. Finally, we included text diacritization as a pre-finetuning task. After each addition, the model was again finetuned on the ATD task to gauge the incremental benefit brought by the new pre-finetuning task. The models used in our ablation study, including the TCO, CA, CA-POS, CA-POS-Seg, and FULL model (PTCAD(0)), are defined and described in subsection \ref{overv}.

\begin{table}[H]
\resizebox{\textwidth}{!}{%
\begin{tabular}{lllllllll}
\textbf{}                                                                                                                                  & \multicolumn{4}{c}{\textbf{Pre-finetuning tasks}}                                  & \multicolumn{2}{c}{\textbf{Abbad Tashkeela}} & \multicolumn{2}{c}{\textbf{Fadel Tashkeela}} \\ \cline{2-9} 
                                                                                                                                           & CA Arabic          & POS Tag            & Segmentation       & Text Diacritization & DER                                            & WER                                            & DER                                           & WER                                          \\ \hline
\textbf{TCO}                                                                                             & \textbf{\ding{55}} & \textbf{\ding{55}} & \textbf{\ding{55}} & \textbf{\ding{55}}  & 0.95                                           & 3.78                                           & 1.50                                          & 6.1                                          \\ \hline
\textbf{CA}                                                                                                  & \textbf{\ding{51}} & \textbf{\ding{55}} & \textbf{\ding{55}} & \textbf{\ding{55}}  & 0.94                                           & 3.75                                           & 1.49                                          & 5.99                                         \\ \hline
\textbf{CA-POS}                                       & \textbf{\ding{51}} & \textbf{\ding{51}} & \textbf{\ding{55}} & \textbf{\ding{55}}  & 0.93                                           & 3.7                                            & 1.44                                          & 5.66                                         \\ \hline
\textbf{CA-POS-Seg}                       & \textbf{\ding{51}} & \textbf{\ding{51}} & \textbf{\ding{51}} & \textbf{\ding{55}}  & 0.93                                           & 3.68                                           & 1.40                                          & 5.51                                         \\ \hline
\textbf{FULL (PTCAD(0))} & \textbf{\ding{51}} & \textbf{\ding{51}} & \textbf{\ding{51}} & \textbf{\ding{51}}  & \textbf{0.89}                                  & \textbf{3.54}                                  & \textbf{1.13}                                 & \textbf{4.27}   \\\hline                            
\end{tabular}
}
\caption{Ablation experiments over each pre-finetuning task to assess its relative importance}
\label{ablation}
\end{table}

The results, meticulously detailed in Table \ref{ablation}, illustrate a clear pattern: each added pre-finetuning task incrementally enhanced the accuracy of our model. This incremental improvement was consistently observed across different evaluation metrics, including DER and WER, on the two benchmarks: Abbad Tashkeela and Fadel Tashkeela. Notably, the most significant improvement was observed when text diacritization was included as a pre-finetuning task, underscoring its pivotal role in enhancing model performance for ATD. This comprehensive analysis reaffirms that each pre-finetuning task contributes significantly to refining our model's ability to handle the nuances of ATD, thereby establishing a strong foundation for our PTCAD system.

\subsection{Error Analysis}

In this subsection, we conduct an in-depth error analysis of our PTCAD(0) system. Despite achieving a notably low error rate on both benchmarks, it is crucial to dissect the nature and origins of these errors. Such scrutiny not only highlights the present limitations of our models but also paves the way for targeted improvements and future research trajectories.
\\

\begin{figure}[h]
\centering
\includegraphics[width=\linewidth]{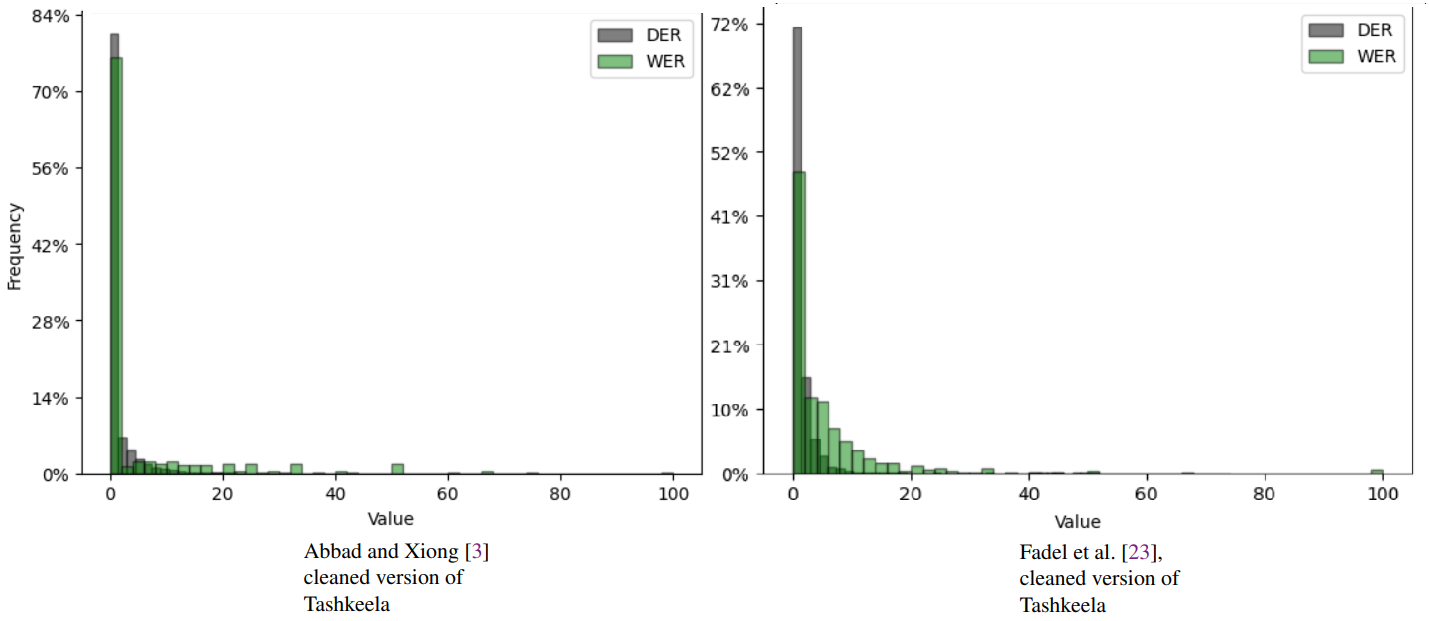}
\caption{Histogram of DER and WER for the Abbad Tashkeela and Fadel Tashkeela benchmarks}
\label{fig:hist_atd_dnn}
\end{figure}

\noindent
\textbf{General Statistics and Error Distribution Patterns.}
Herein, we provide detailed statistics concerning the distribution of DER and WER metrics across our two benchmarks. Key metrics encompass the count of sentences with a DER and WER of zero, along with a breakdown of DER across different ranges.\\
Figure \ref{fig:hist_atd_dnn} depicts a histogram of DER and WER for the two benchmark datasets Abbad Tashkeela and Fadel Tashkeela. This visual representation elucidates the frequency distribution of errors, revealing the prevalence of various error rates within our test dataset.

\noindent
\textbf{Quantitative Overview of Error Distribution.}
To objectively evaluate the error spread, table \ref{stats_error} encapsulates crucial statistics reflecting our model's accuracy. This data aims to detail the proportion of test sentences with no errors and categorize the spread of DER and WER across defined ranges [Min Error Rate, Max Error Rate].

\begin{table}[H]
\resizebox{\textwidth}{!}{%
\centering
\begin{tabular}{l|ccc|ccc}
\hline
\multirow{2}{*}{\textbf{Benchmark}} & \multicolumn{3}{c|}{\textbf{Abbad Tashkeela}}                 & \multicolumn{3}{c}{\textbf{Fadel Tashkeela}}                    \\ \cline{2-7} 
                                    & \multicolumn{1}{c|}{Min Error Rate} & \multicolumn{1}{c|}{Max Error Rate} & Proportion of sentences (\%) & \multicolumn{1}{c|}{Min Error Rate} & \multicolumn{1}{c|}{Max Error Rate} & Proportion of sentences (\%) \\ \hline
WER                                 & \multicolumn{1}{c|}{0}   & \multicolumn{1}{c|}{0}   & 76         & \multicolumn{1}{c|}{0}   & \multicolumn{1}{c|}{0}   & 72         \\
                                    & \multicolumn{1}{c|}{0}   & \multicolumn{1}{c|}{30}  & 20         & \multicolumn{1}{c|}{0}   & \multicolumn{1}{c|}{20}  & 22         \\
                                    & \multicolumn{1}{c|}{30}  & \multicolumn{1}{c|}{100} & 4          & \multicolumn{1}{c|}{20}  & \multicolumn{1}{c|}{100} & 6          \\ \hline
DER                                 & \multicolumn{1}{c|}{0}   & \multicolumn{1}{c|}{0}   & 76         & \multicolumn{1}{c|}{0}   & \multicolumn{1}{c|}{0}   & 72         \\
                                    & \multicolumn{1}{c|}{0}   & \multicolumn{1}{c|}{10}  & 22         & \multicolumn{1}{c|}{0}   & \multicolumn{1}{c|}{5}   & 22         \\
                                    & \multicolumn{1}{c|}{10}  & \multicolumn{1}{c|}{100} & 2          & \multicolumn{1}{c|}{5}   & \multicolumn{1}{c|}{50}  & 6    \\\hline     
\end{tabular}
}
\caption{Statistical overview of PTCAD performance}
\label{stats_error}
\end{table}
For each metric and benchmark, we distinguish three classes based on the error rates:
\begin{itemize}
    \item Class with the Lowest Error Rate (0\%): This class is predominant, accounting for approximately 76\%.
    \item Medium Error Class: For DER, this class ranges between 0 and 10\%, and for WER, it is between 0 and 30\%. This class represents roughly 20\% of the cases.
    \item High Error Class: This class includes cases with more than 10\% DER or more than 30\% WER. It comprises a smaller proportion, approximately 5\% of the cases.
\end{itemize}
These statistics and graphical analyses establish a fundamental comprehension of the model's error landscape. This forms the basis for subsequent sections that delve into the specific causes and characteristics of these errors.
\\

\noindent
\textbf{In-Depth Error Analysis.} To gain targeted insights into the model's weaknesses, we specifically analyzed  1000 sentences from Abbad Tashkeela and 200 sentences from Fadel Tashkeela where the model performed the poorest (minimum error rate greater than 30\%), aiming to understand the types and causes of errors in these instances.
\\
\\
\textit{Fact 1. Effect of Sentence Length.} A notable observation from our analysis was the impact of sentence length on diacritization accuracy:

\begin{itemize}
    \item In Abbad Tashkeela, over 80\% of the sentences had 5 words or fewer.
    \item In Fadel Tashkeela, this figure was approximately 33\%.
\end{itemize}
Short sentences pose a particular challenge for diacritization due to their limited contextual information. The paucity of context can lead to multiple correct diacritization possibilities, making it difficult for the model to match the exact ground truth. Consequently, these short sentences were excluded from our detailed analysis, focusing instead on sentences with more than 5 words.
\\
\\
\textit{Fact 2. Effect of Sentence Truncation on the Performance of PTCAD(0).}
Another key factor we considered was sentence truncation due to our model's maximum token sequence length of 512. If a sentence exceeded this limit and was split, it could potentially lose vital context, impacting the diacritization performance. This was particularly relevant for longer sentences that are substrings of original, longer sentences.
\begin{itemize}
    \item For Abbad Tashkeela, less than 10 sentences among the analyzed 1000, accounting for 1\%, were affected by truncation.
    \item For Fadel Tashkeela, 16 sentences, accounting for 8\%, were truncated.
\end{itemize}
All truncated sentences showed direct error causation from the cut, especially evident in the initial words post-truncation and, occasionally, further into the sentence.
\\
\\
Example 1:\\
\textit{Original sentence (not fully displayed):}
\setcode{utf8}
\begin{eqnarray*}
& \RL{[...] وَأَمَّا نَحْوُ الْقِرَاءَةِ فِي سَبْعٍ أَوْ أَجْزَاءٍ يُقَصِّرُ الْوَاحِدُ مِنْهُمْ أَحْيَانَا أَوْ يَمُوتُ ، فَإِنْ كَانُوا مُعَيَّنِينَ فَهُمْ }\\
& \RL{كَالْأَجْزَاءِ لِكُلٍّ وَاحِدٌ أَوْ وَارِثُهُ بِقَدْرِ عَمَلِهِ وَإِلَّا لَمْ يُعْطَ شَيْئًا وَاَللَّهُ أَعْلَمُ .( 6 / 418 ) }\\
\end{eqnarray*}
\textit{Sentence cut to diacritize:}
$$\RL{ لِكُلٍّ وَاحِدٌ أَوْ وَارِثُهُ بِقَدْرِ عَمَلِهِ وَإِلَّا لَمْ يُعْطَ شَيْئًا وَاَللَّهُ أَعْلَمُ .( 6 / 418 )}$$
\textit{Model Prediction:}
$$\RL{لِكُلِّ وَاحِدٍ أَوْ وَارِثِهِ بِقَدْرِ عَمَلِهِ وَإِلَّا لَمْ يُعْطِ شَيْئًا وَاَللَّهُ أَعْلَمُ .( 6 / 418 )}$$
\textit{Error for this sentence:}
\begin{eqnarray*}
    DER &=& 7.35\\
    WER &=& 33.33
\end{eqnarray*}
Example 2:\\
\textit{Original sentence (not fully displayed):}
\setcode{utf8}
\begin{eqnarray*}
& \RL{[...]ثُمَّ قَالَ : وَفِي هَذَا تَضْيِيقٌ شَدِيدٌ وَعَمَلُ النَّاسِ عَلَى خِلَافِهِ مِنْ غَيْرِ نَكِيرٍ وَعَلَى الْأَوَّلِ الْأَوْجَهِ }\\
& \RL{  فَهَلْ الْمُرَادُ بِالْمَحِلِّ فِي كَلَامِهِ الْمَحَلَّةُ الَّتِي هُوَ فِيهَا كَنَقْلِ الزَّكَاةِ أَوْ مَوْضِعُهُ}\\
& \RL{   الْمَنْسُوبُ إلَيْهِ عَادَةً بِحَيْثُ يَقْصِدُ الْمُسَبَّلَ أَهْلُهُ بِذَلِكَ ؟ مَحِلُّ نَظَرٍ وَالثَّانِي أَقْرَبُ ،}\\
\end{eqnarray*}

\textit{Sentence cut to diacritize:}
\begin{eqnarray*}
& \RL{الْأَوَّلِ الْأَوْجَهِ فَهَلْ الْمُرَادُ بِالْمَحِلِّ فِي كَلَامِهِ الْمَحَلَّةُ الَّتِي هُوَ فِيهَا كَنَقْلِ الزَّكَاةِ أَوْ مَوْضِعُهُ الْمَنْسُوبُ إلَيْهِ}\\
& \RL{عَادَةً بِحَيْثُ يَقْصِدُ الْمُسَبَّلَ أَهْلُهُ بِذَلِكَ ؟ مَحِلُّ نَظَرٍ وَالثَّانِي أَقْرَبُ ، }\\
\end{eqnarray*}
\textit{Model Prediction:}
\begin{eqnarray*}
& \RL{الْأَوَّلُ الْأَوْجَهُ فَهَلْ الْمُرَادُ بِالْمَحَلِّ فِي كَلَامِهِ الْمَحَلَّةُ الَّتِي هُوَ فِيهَا كَنَقْلِ الزَّكَاةِ أَوْ مَوْضِعِهِ الْمَنْسُوبِ إلَيْهِ }\\
& \RL{عَادَةً بِحَيْثُ يَقْصِدُ الْمُسَبِّلُ أَهْلَهُ بِذَلِكَ ؟ مَحَلُّ نَظَرٍ وَالثَّانِي أَقْرَبُ ،}\\
\end{eqnarray*}
\textit{Error for this sentence:}
\begin{eqnarray*}
    DER &=& 5.75\\
    WER &=& 27.59
\end{eqnarray*}
To address the issue of sentence truncation, as previously mentioned in the subsection \ref{benchmarking_subsec}, we have developed a second strategy in addition to \textbf{PTCAD(0)}: the sliding window \textbf{PTCAD(p)}. This strategy involves a careful sliding of the window by 'p' words to maintain context while diacritizing. Our experiments revealed that a sliding window size of 'p=5' words offered the best performance for this method. For the Fadel Tashkeela benchmark, we observed a notable impact of this strategy:
\begin{itemize}
    \item On long sentences, the performance before implementing \textbf{PTCAD(p)} was a WER of 4\% and a DER of 1.06\% using just PTCAD(0). After applying the \textbf{PTCAD(p)} strategy, the performance improved to a DER of 1\% and a WER of 3.77\%.
    \item Across the entire set of sentences, the initial performance was a DER of 1.13\% and a WER of 4.27\%. After implementation, the performance slightly enhanced to a DER of 1.1\% and a WER of 4.19\%.
\end{itemize}
Similarly, for the Abbad Tashkeela benchmark:
\begin{itemize}
    \item On long sentences, the initial performance was a WER of 3.38\% and a DER of 0.89\%. Post \textbf{PTCAD(p)} implementation, there was a significant improvement with the DER dropping to 0.77\% and the WER to 2.89\%.
    \item For the entire sentence set, the performance improved from an initial DER of 0.89\% and a WER of 3.54\% to a DER of 0.87\% and a WER of 3.53\% after applying \textbf{PTCAD(p)}.
\end{itemize}
These results underscore the effectiveness of the sliding window approach in addressing the challenges posed by sentence truncation in our PTCAD model, particularly for longer sentences. The improvements in both DER and WER metrics highlight the strategy's capability to preserve the contextual integrity of sentences, thereby enhancing overall diacritization accuracy.
\\
\\
\textit{Fact 3. Categories of Errors Identified.}
In addition to exploring the unique category of errors stemming from sentence truncation, our investigation also identified a range of other error types. We have strived for thoroughness in this analysis. However, it's important to note that the percentages in the table \ref{category_error}, indicating the proportions of each error type, are rough estimates. This is due to the labor-intensive process of manually reviewing and categorizing each error type in the test sentences. The table outlines these error categories and their estimated occurrence proportions in the analyzed sentences from both benchmarks, but it might not capture every possible error scenario.

\begin{table}[H]
\centering
\resizebox{410pt}{!}{
\begin{tabular}{l|l|cc|l}
\textbf{Error Category }                                                                 & \textbf{Description}                                                                                                                                                                                                                                                                                                                                                                                                                            & \multicolumn{1}{l|}{\begin{tabular}[c]{@{}l@{}}\textbf{Estimated}\\ \textbf{Proportion}\\ \textbf{Abbad} \\\textbf{Tashkeela (\%)}\end{tabular}} & \multicolumn{1}{l|}{\begin{tabular}[c]{@{}l@{}}\textbf{Estimated}\\ \textbf{Proportion}\\ \textbf{Fadel} \\\textbf{Tashkeela (\%)}\end{tabular}} & \textbf{Examples}                                                                                                                                                                                                                                                                                                                                                                                                                                             \\ \hline
\textbf{Limited Context  }                                                               & \begin{tabular}[c]{@{}l@{}}Errors due to insufficient context, especially in shorter sentences\end{tabular}                                                                                                                                                                                                                                                                                                                          & \multicolumn{1}{c|}{30}                                                                      & 40                                                                                           & $$[...]\RL{ فَقَالَ ابْنُ الْكَويِ : دَعُوهُمْ }\textcolor{green}{\RL{وَيُلَمُّ شَعْثُهَا}}\sout{\textcolor{red}{\RL{وَيَلُمُّ شَعَثَهَا}}}$$                                                                                                                                                                                                                                                                                                        \\ \hline
\begin{tabular}[c]{@{}l@{}}\textbf{Ambiguity in}\\ \textbf{Diacritization}\end{tabular}           & \begin{tabular}[c]{@{}l@{}}Discrepancies due to multiple valid diacritization forms\end{tabular}                                                                                                                                                                                                                                                                                                                                     & \multicolumn{2}{c|}{40}                                                                                                                                                                     & $$\textcolor{green}{\RL{رِهِ}}\sout{\textcolor{red}{\RL{رُهُ}}}\RL{ وَقَدْ}\textcolor{green}{\RL{طِهِ}}\sout{\textcolor{red}{\RL{طُهُ}}}\RL{ وَشَرْ}\textcolor{green}{\RL{نِهِ}}\sout{\textcolor{red}{\RL{نُهُ}}}\RL{ وَرُكْ}\textcolor{green}{\RL{بِهِ}}\sout{\textcolor{red}{\RL{بُهُ}}}\RL{ وَسَبَ}$$                                                                                                                                             \\ \hline
\begin{tabular}[c]{@{}l@{}}\textbf{Sentence}\\ \textbf{Truncation}\end{tabular}                   & \begin{tabular}[c]{@{}l@{}}Errors from context loss in truncated sentences\end{tabular}                                                                                                                                                                                                                                                                                                                                              & \multicolumn{1}{c|}{8}                                                                       & 0                                                                                            & $$\RL{ يَحْرُمُ لِلضَّرَرِ كَمَا مَرَّ }\RL{يَّةُ}\textcolor{green}{\RL{مِّ}}\sout{\textcolor{red}{\RL{مُ}}}\RL{ السُّ}\RL{الْبَحْرِ الَّذِي فِيهِ }\textcolor{green}{\RL{نَ}}\sout{\textcolor{red}{\RL{نُ}}}\RL{ حَيَوَا}[...]$$                                                                                                                                                                                                              \\ \hline
\begin{tabular}[c]{@{}l@{}}\textbf{Complex}\\ \textbf{Language Forms}\end{tabular}                & \begin{tabular}[c]{@{}l@{}}Struggles with rare words, complex structures, and metaphorical\\ language\end{tabular}                                                                                                                                                                                                                                                                                                                     & \multicolumn{1}{c|}{14}                                                                      & 10                                                                                           & \begin{tabular}[c]{@{}l@{}}Rare words: $\RL{بِمَكُّوكٍ}$, $\RL{السُّمُيَّةُ}$\\ Complex sentences:\\ $$\RL{يٌّ ذَاتِيٌّ }\textcolor{green}{\RL{عَرِ}}\sout{\textcolor{red}{\RL{عُرْ}}}\RL{ فِي المِرْآةِ الوَرَقِيَّةِ }$$\end{tabular}                                                                                                                                                                                                              \\ \hline
\begin{tabular}[c]{@{}l@{}}\textbf{Benchmark}\\ \textbf{Inaccuracies}\end{tabular}                & Errors in the benchmark datasets                                                                                                                                                                                                                                                                                                                                                                                                       & \multicolumn{1}{c|}{2}                                                                       & 5                                                                                            & \begin{tabular}[c]{@{}l@{}}Errors: $\RL{ أَصْلًا }\textcolor{red}{\RL{رُهُ}}\RL{شَيْءٌ عَلَى غَيْ }$\\ Errors: $\RL{طْبَةٍ}\textcolor{red}{\RL{خِ}}\RL{عَةُ إِلَّا بِ}\textcolor{red}{\RL{مْ}}\RL{لَا تَكُونُ الجُ }$\\ Incomplete diacritization: \\ $$\RL{صْحَ مَقْبُولُ}\textcolor{red}{\RL{إنّ النّ }}\RL{هَا صَدَقَتْ ... بِوَعْدِهَا أَوْ لَوْ}\textcolor{red}{\RL{نّ}}\RL{ةٌ لَوْ أَ}\textcolor{red}{\RL{لّ}}\RL{فَيَا لَهَا حُ }$$\end{tabular} \\ \hline
\begin{tabular}[c]{@{}l@{}}\textbf{Independent}\\ \textbf{Token}\\ \textbf{Classification}\end{tabular}    & \begin{tabular}[c]{@{}l@{}}These errors occur when PTCAD systems, focusing on individual \\ token diacritization, produce contextually inconsistent results. This \\issue becomes prominent in sentences where multiple diacritizations \\ are plausible, leading to a mix of correct yet collectively incoherent \\diacritical marks.\end{tabular} & \multicolumn{2}{c|}{2}                                                                                                                                                                      & $$\RL{هَا}\textcolor{green}{\RL{ثَ}}\sout{\textcolor{red}{\RL{ثُ}}}\RL{سَوَالِفَ زَلَّاتِي و حَوَادِ}$$                                                                                                                                                                                                                                                                                                                                              \\ \hline
\begin{tabular}[c]{@{}l@{}}\textbf{Overlooking}\\ \textbf{Nuanced}\\ \textbf{Linguistic Cues}\end{tabular} & \begin{tabular}[c]{@{}l@{}}The model misses subtle linguistic hints crucial for correct\\ diacritization. It often ignores key phrases that guide proper diacritical\\ marking, resulting in contextually incorrect interpretations.\end{tabular}                                                                                                                                                                                  & \multicolumn{2}{c|}{1}                                                                                                                                                                      & $$[...]\RL{زَّةً بِكَسْرِهِمَا }\textcolor{green}{\RL{عِ}}\sout{\textcolor{red}{\RL{عَ}}}\RL{زًّا وَ}\textcolor{green}{\RL{عِ}}\sout{\textcolor{red}{\RL{عَ}}}\RL{وَيُقَالُ عَزَّ }$$                                                                                                                                                                                                                                                                \\ \hline
\begin{tabular}[c]{@{}l@{}}\textbf{Unexplained}\\ \textbf{Errors}\end{tabular}                    & \begin{tabular}[c]{@{}l@{}}Errors not fitting into any specific category, indicating unknown factors\end{tabular}                                                                                                                                                                                                                                                                                                                    & \multicolumn{2}{c|}{3}                                                                                                                                                                      & $\RL{الرّجُلُ}$, $\RL{بَاللُهجةُ}$, ...                                                                                                                                                                                                                                                                                                                                                                                                              \\ \hline
\end{tabular}
}
\caption{Categorization and proportion of identified errors in PTCAD system}
\label{category_error}
\end{table}

This error analysis reveals crucial insights into the challenges faced by our ATD system. Addressing issues related to context insufficiency, sentence truncation, and language complexity is essential for improving the model. Enhancing the training data and refining the model's ability to understand and adapt to different language nuances will be central to our future efforts, aiming for a more robust and accurate ATD system.

\section{Advantages and Limitations of our approach}

\subsection{Advantages}

\subsubsection*{Computational Efficiency}
Our model harnesses the computational efficiency characteristic of BERT-like architectures, making it uniquely capable of conducting inference operations smoothly even on CPUs. This aspect greatly extends the versatility and practicality of our approach across various computing environments.

\subsubsection*{Robustness of BERT}
Our model inherits all the robust features of BERT-like architectures, such as excellent contextual understanding and generalization capabilities. This is particularly beneficial for a language as contextually rich and nuanced as Arabic.

\subsubsection*{Consistent Number of Diacritics}
Unlike previous models, our approach ensures a fixed number of diacritics for each word, based on the number of Arabic letters it contains. This constrains the search space to a set of 14 diacritics, thereby increasing prediction accuracy.

\subsubsection*{Ease of Reproduction}
Our approach simplifies the diacritization task into a token classification problem, making the reproduction of this model straightforward. One only needs to prepare the data and utilize a pre-trained BERT model for token classification.

\subsubsection*{Limited Contextual Ambiguity}
While the model does have limitations in dealing with contextually ambiguous situations, these are relatively rare. When the context is sufficiently informative, the model performs exceptionally well in diacritizing the text.

\subsection{Limitations}

\subsubsection*{Limitations in Dataset Evaluation}
Our assessment of the PTCAD method did not include tests on additional datasets mentioned in the "Related Work" section (Section \ref{Related Work}). This decision was based on specific constraints associated with each dataset:
\begin{enumerate}
    \item Holy Quran and Sahih al Bukhary. The training and testing data partitioning specified in Lamad's study (\citep{lamad}) is not available, which precludes a standardized evaluation.
    \item The MSA corpus of 4.5 million words \citep{highly_eff} is not publicly available, thus limiting our ability to conduct an assessment.
    \item The Penn Arabic Treebank \citep{patb} is not freely available, posing a financial barrier to its inclusion in our evaluation.
\end{enumerate}

\subsubsection*{Limitations in Test Benchmarks}
Our model evaluation's accuracy is impacted by the benchmark quality, particularly due to inaccuracies in their ground truths. This highlights the need for more refined benchmarks in future research:

\begin{itemize}
    \item \textbf{Accuracy of Benchmarks.} The presence of errors or inaccuracies in current benchmarks may compromise the model's performance assessment. Ideal benchmarks should be error-free, ensuring a high standard for evaluation.
    
    \item \textbf{Completeness of Diacritization.} Current benchmarks often lack full diacritization, presenting partial or inconsistent text. This limits thorough model evaluation across diverse sentence structures. Fully diacritized benchmarks are essential for a comprehensive assessment.
    
    \item \textbf{Contextual Richness of Texts.} Benchmarks predominantly feature short, isolated sentences, lacking comprehensive context. This restricts the model's ability to be evaluated on longer, more coherent texts. Inclusion of extended, context-rich texts in benchmarks is crucial for testing the model's diacritization consistency and accuracy in realistic scenarios.
\end{itemize}

\subsubsection*{Limited Ability to Generalize to Unseen or Complex Language Formats}

One of the limitations of our PTCAD system revealed by the Error Analysis section is the ability to generalize to new or complex language formats. This limitation is in fact primarily tied to the training data's quality and diversity. Training the PTCAD system with a varied dataset of well-diacritized religious texts significantly improved its performance on complex linguistic structures. This richer dataset, with a wide range of language patterns from religious literature, proved key in enhancing the model's adaptability. The substantial progress achieved with this improved dataset underscores the need for a diverse and comprehensive training corpus. Addressing the challenges in processing complex language formats requires a dataset that not only expands in size but also includes varied linguistic complexities and stylistic nuances. Such strategic enrichment of training material prepares the model to more effectively handle unfamiliar or sophisticated linguistic contexts, thereby minimizing errors.

\subsubsection*{Sequence Length and Computational Cost Trade-off}
In addressing long sentence diacritization in our PTCAD model (refer to \ref{benchmarking_subsec}), we face a trade-off between sequence length management and computational efficiency. Our strategies, PTCAD(0) for reduced computation and PTCAD(p) for sliding window processing have distinct impacts. While PTCAD(p) effectively overcomes the sequence length limitation, it significantly increases computational costs due to the repeated processing of overlapping windows. This scenario highlights a critical challenge: balancing the efficient handling of long sentences with the increased computational requirements of these inference strategies.

\subsubsection*{Independent Token Classification and Its Impact on Word-level and Sentence-level Coherence}

BERT-like models are known for their robust contextual understanding. However, in token classification tasks, these models treat each token independently. This approach can result in diacritization that is inconsistent or contextually inappropriate, especially in cases where the sentence context is ambiguous or provides limited cues.
\begin{itemize}
    \item  \textbf{Impact on Word-level Coherence.} 
Take, for example, the sentence $\RL{علمته حقا.}$ The word $\RL{علمته}$ could be diacritized in multiple ways: $\RL{عَلَّمَتْهُ}$ (she taught it to him) or $\RL{عَلِمْتَهُ}$ (you knew it). Without sufficient context, both diacritizations could be considered correct. However, if diacritical marks are predicted independently, it might lead to an inconsistent output like $\RL{عَلَّمَتَهُ}$. In this scenario, the first three diacritical marks suggest $\RL{عَلَّمَتْهُ}$, while the last one aligns with $\RL{عَلِمْتَهُ}$, resulting in a meaningless amalgamation.

\item \textbf{Impact on Sentence-level Coherence.} 
While these models may diacritize individual words accurately, maintaining coherence at the sentence level remains a significant challenge. Instances occur where the diacritization is accurate on a word-by-word basis but becomes illogical when the words are combined into a sentence. For instance: $$\RL{هَا}\textcolor{green}{\RL{ثَ}}\textcolor{red}{\sout{{\RL{ثُ}}}}\RL{سَوَالِفَ زَلَّاتِي و حَوَادِ}$$
\end{itemize}

\section{Conclusion and Future Work}

In this study, we have presented a novel approach to the Automatic Diacritization of Arabic text, leveraging the robust capabilities of pre-trained BERT-like models. Our methodology innovatively combines the strengths of these models with a new 'pre-finetuning' phase, incorporating linguistically relevant tasks like POS tagging, text segmentation, and CA finetuning. Our approach was rigorously tested across two benchmark datasets, achieving a significant 20\% reduction in WER and setting a new state-of-the-art. This work not only advances the field of ATD but also provides a blueprint for employing pre-trained language models in highly specialized NLP tasks. Despite its strengths, the study also uncovers several limitations, such as issues with sequence length and potential inconsistencies in diacritization due to treating tokens independently. Thus, several exciting avenues for future work have been identified:
\\
\\
\textbf{In terms of models and architectures.}
\begin{itemize}
    \item \textit{Improving Token Consistency.} One could investigate adding an LSTM layer on top of the token classification head over each separate sequence of masks. This modification aims to ensure that the predicted diacritics are consistent and contextually appropriate.
    
    \item \textit{Generative Models.} Another promising direction is the exploration of generative models like LLama2 for ATD. These models can be finetuned on a diverse set of related tasks to gain a comprehensive understanding of Arabic language dynamics. The enriched context provided by these models could offer a different approach to predicting diacritics and might further improve accuracy.
    
    \item \textit{Sentence-level Coherence.} Future work could also focus on extending the model's capabilities to ensure sentence-level coherence in diacritization, which remains an open challenge.
    
    \item \textit{Computational Efficiency.} Further optimization techniques could be explored to make the model even more efficient, allowing it to handle longer sequences without compromising on the quality of diacritization.
\end{itemize}

\noindent
\textbf{In terms of training data.} The limitations in generalizing to new or complex language formats, identified in the Error Analysis, are primarily attributed to the training data's quality and diversity. In our experimentation, we have explored enhancing the PTCAD system with a richer dataset, including well-diacritized religious texts. Preliminary results from these experiments suggest potential improvements in the model's ability to handle complex linguistic structures. This exploration indicates that a varied dataset with intricate language patterns could be key to increasing model adaptability.

\begin{itemize}
    \item \textit{Rich Linguistic Formats:} Future endeavors should focus on incorporating training data that encompasses a broader array of linguistic styles and complexities.

    \item \textit{Enhancing Sensitivity to Linguistic Cues:} The model's current shortfall in recognizing subtle linguistic cues crucial for accurate diacritization needs addressing. Training should include sentences that embed key phrases and nuanced patterns, helping the model discern contextually vital diacritical marks. This strategy aims to refine its interpretative accuracy, exemplified by complex sentences like $$[...]\RL{زَّةً بِكَسْرِهِمَا }\textcolor{green}{\RL{عِ}}\sout{\textcolor{red}{\RL{عَ}}}\RL{زًّا وَ}\textcolor{green}{\RL{عِ}}\sout{\textcolor{red}{\RL{عَ}}}\RL{وَيُقَالُ عَزَّ }$$.
\end{itemize}

\noindent
\textbf{In terms of test benchmark.} Developing robust benchmarks is essential for accurately assessing the model's performance and its applicability in real-world scenarios.

\begin{itemize}
    \item \textit{Error-Free Benchmarks.} Establishing a benchmark with completely accurate and error-free texts will provide a clear standard for model evaluation. This ensures that the model's performance is measured against the highest accuracy standards.

    \item \textit{Comprehensive Diacritization.} Benchmarks should consist of fully diacritized sentences, avoiding partial or incomplete diacritization. This will challenge the model to recognize and apply diacritics across a wide range of sentence structures and contexts.

    \item \textit{Context-Rich Texts.} The benchmarks should include texts that offer full context, moving away from short, isolated sentences to more coherent and extended texts. This will test the model's ability to maintain consistency and accuracy in diacritization over longer passages, reflecting more realistic usage scenarios.
\end{itemize}

The methodologies and findings of this research set the stage for subsequent studies that can push the boundaries of what is achievable in ATD. The strategies outlined for future work have the potential to resolve existing limitations and to take the field a step closer to the ultimate goal of perfectly diacritized Arabic text.

\bibliography{compling_style}

\begin{thebibliography}{}

\bibitem[Abandah et~al., 2022a]{classif_diac}
Abandah, G.~A., Khedher, M.~Z., Abdel-Majeed, M.~R., Mansour, H.~M., Hulliel, S.~F., and Bisharat, L.~M. (2022a).
\newblock Classifying and diacritizing arabic poems using deep recurrent neural networks.
\newblock {\em Journal of King Saud University-Computer and Information Sciences}, 34(6):3775--3788.

\bibitem[Abandah et~al., 2022b]{transfer_ad_poet}
Abandah, G.~A., Suyyagh, A.~E., and Abdel-Majeed, M.~R. (2022b).
\newblock Transfer learning and multi-phase training for accurate diacritization of arabic poetry.
\newblock {\em Journal of King Saud University-Computer and Information Sciences}, 34(6):3744--3757.

\bibitem[Abbad and Xiong, 2020]{multi_comp}
Abbad, H. and Xiong, S. (2020).
\newblock Multi-components system for automatic arabic diacritization.
\newblock In {\em Advances in Information Retrieval: 42nd European Conference on IR Research, ECIR 2020, Lisbon, Portugal, April 14--17, 2020, Proceedings, Part I 42}, pages 341--355. Springer.

\bibitem[Abbad and Xiong, 2021]{simple_ext}
Abbad, H. and Xiong, S. (2021).
\newblock Simple extensible deep learning model for automatic arabic diacritization.
\newblock {\em Transactions on Asian and Low-Resource Language Information Processing}, 21(2):1--16.

\bibitem[Abdul-Mageed et~al., 2012]{abdul2012arabic}
Abdul-Mageed, M., Diab, M., and Rambow, O. (2012).
\newblock Arabic diacritization using conditional random fields.
\newblock {\em Computational Linguistics}, 38(3):585--612.

\bibitem[Abdul-Mageed et~al., 2021]{arbert}
Abdul-Mageed, M., Elmadany, A., and Nagoudi, E. M.~B. (2021).
\newblock Arbert \& marbert: Deep bidirectional transformers for arabic.

\bibitem[Al et~al., 2022]{custo}
Al, A., Rahman, M., Islam, M., and et~al. (2022).
\newblock Natural language processing in customer service: A systematic review.
\newblock {\em arXiv preprint arXiv:2212.09523}.

\bibitem[Al-Hajj and Al-Rawi, 2004]{alhajj2004arabic}
Al-Hajj, S. and Al-Rawi, K. (2004).
\newblock Arabic diacritization: A survey and some improvements.
\newblock {\em Journal of the American Society for Information Science and Technology}, 55(6):495--505.

\bibitem[Al-Sabri and Gao, 2021]{lamad}
Al-Sabri, R. and Gao, J. (2021).
\newblock Lamad: A linguistic attentional model for arabic text diacritization.
\newblock In {\em Findings of the Association for Computational Linguistics: EMNLP 2021}, pages 3757--3764.

\bibitem[AlKhamissi et~al., 2020]{efficient_hier}
AlKhamissi, B., ElNokrashy, M.~N., and Gabr, M. (2020).
\newblock Deep diacritization: Efficient hierarchical recurrence for improved arabic diacritization.
\newblock In {\em Workshop on Arabic Natural Language Processing}.

\bibitem[Almanea, 2021]{survey}
Almanea, M.~M. (2021).
\newblock Automatic methods and neural networks in arabic texts diacritization: A comprehensive survey.
\newblock {\em IEEE Access}, 9:145012--145032.

\bibitem[Alqahtani et~al., 2020]{multi_task_ad}
Alqahtani, S., Mishra, A., and Diab, M.~T. (2020).
\newblock A multitask learning approach for diacritic restoration.
\newblock In Jurafsky, D., Chai, J., Schluter, N., and Tetreault, J.~R., editors, {\em Proceedings of the 58th Annual Meeting of the Association for Computational Linguistics, {ACL} 2020, Online, July 5-10, 2020}, pages 8238--8247. Association for Computational Linguistics.

\bibitem[Antoun et~al., 2020]{arabert}
Antoun, W., Baly, F., and Hajj, H. (2020).
\newblock Arabert: Transformer-based model for arabic language understanding.
\newblock In {\em LREC 2020 Workshop Language Resources and Evaluation Conference 11--16 May 2020}, page~9.

\bibitem[Anwar, 2018]{tashkeelamodel}
Anwar, M. (2018).
\newblock Tashkeela-model.
\newblock https://github.com/Anwarvic/TashkeelaModel.

\bibitem[Barqawi and Zerrouki, 2017]{shakkala}
Barqawi, A. and Zerrouki, T. (2017).
\newblock Shakkala, arabic text vocalization.
\newblock {\em Retrieved February}, 20:2021.

\bibitem[Belinkov et~al., 2019]{openiti}
Belinkov, Y., Magidow, A., Barrón-Cedeño, A., Shmidman, A., and Romanov, M. (2019).
\newblock Studying the history of the arabic language: language technology and a large-scale historical corpus.
\newblock {\em Language Resources and Evaluation}, 53.

\bibitem[Beltagy et~al., 2020]{Longformer}
Beltagy, I., Peters, M.~E., and Cohan, A. (2020).
\newblock Longformer: The long-document transformer.
\newblock {\em CoRR}, abs/2004.05150.

\bibitem[Darwish et~al., 2021]{ad_feature_rich}
Darwish, K., Abdelali, A., Mubarak, H., and Eldesouki, M. (2021).
\newblock Arabic diacritic recovery using a feature-rich bilstm model.
\newblock {\em Transactions on Asian and Low-Resource Language Information Processing}, 20(2):1--18.

\bibitem[Darwish et~al., 2017]{darwish-etal-2017}
Darwish, K., Mubarak, H., and Abdelali, A. (2017).
\newblock {A}rabic diacritization: Stats, rules, and hacks.
\newblock In {\em Proceedings of the Third {A}rabic Natural Language Processing Workshop}, pages 9--17, Valencia, Spain. Association for Computational Linguistics.

\bibitem[Diab et~al., 2013]{diab2013}
Diab, M., Habash, N., Rambow, O., and Roth, R. (2013).
\newblock Ldc arabic treebanks and associated corpora: Data divisions manual.

\bibitem[Ding et~al., 2020]{Ding2020CogLTXAB}
Ding, M., Zhou, C., Yang, H., and Tang, J. (2020).
\newblock Cogltx: Applying bert to long texts.
\newblock In {\em Neural Information Processing Systems}.

\bibitem[El~Mekki et~al., 2020]{el2020weighted}
El~Mekki, A., Alami, A., Alami, H., Khoumsi, A., and Berrada, I. (2020).
\newblock Weighted combination of bert and n-gram features for nuanced arabic dialect identification.
\newblock In {\em Proceedings of the Fifth Arabic Natural Language Processing Workshop}, pages 268--274.

\bibitem[El~Mekki et~al., 2021a]{el2021domain}
El~Mekki, A., El~Mahdaouy, A., Berrada, I., and Khoumsi, A. (2021a).
\newblock Domain adaptation for arabic cross-domain and cross-dialect sentiment analysis from contextualized word embedding.
\newblock In {\em Proceedings of the 2021 conference of the North American chapter of the Association for Computational Linguistics: Human Language Technologies}, pages 2824--2837.

\bibitem[El~Mekki et~al., 2021b]{el-mekki-etal-2021-bert}
El~Mekki, A., El~Mahdaouy, A., Essefar, K., El~Mamoun, N., Berrada, I., and Khoumsi, A. (2021b).
\newblock {BERT}-based multi-task model for country and province level {MSA} and dialectal {A}rabic identification.
\newblock In {\em Proceedings of the Sixth Arabic Natural Language Processing Workshop}, pages 271--275, Kyiv, Ukraine (Virtual). Association for Computational Linguistics.

\bibitem[{European Language Resources Association}, 2001]{annahar}
{European Language Resources Association} (2001).
\newblock An-nahar newspaper text corpus.
\newblock http://www.language-archives.org/item/oai:catalogue.elra.info:ELRA-W0027.

\bibitem[Face, 2023]{training_args}
Face, H. (2023).
\newblock Transformers training args.
\newblock https://github.com/huggingface/transformers/blob/main/src/transformers/training\_args.py.

\bibitem[Fadel et~al., 2019a]{atd_dnn}
Fadel, A., Tuffaha, I., Al-Jawarneh, B., and Al-Ayyoub, M. (2019a).
\newblock Arabic text diacritization using deep neural networks.
\newblock In {\em 2019 2nd International Conference on Computer Applications \& Information Security (ICCAIS)}, pages 1--7.

\bibitem[Fadel et~al., 2019b]{fadel-etal-2019-neural}
Fadel, A., Tuffaha, I., Al-Jawarneh, B., and Al-Ayyoub, M. (2019b).
\newblock Neural {A}rabic text diacritization: State of the art results and a novel approach for machine translation.
\newblock In {\em Proceedings of the 6th Workshop on Asian Translation}, pages 215--225, Hong Kong, China. Association for Computational Linguistics.

\bibitem[Gheith~Abandah, 2020]{abanda}
Gheith~Abandah, A. A.-K. (2020).
\newblock Accurate and fast recurrent neural network solution for the automatic diacritization of arabic text.
\newblock {\em Jordanian Journal of Computers and Information Technology (JJCIT)}, 06(02):103 -- 121.

\bibitem[Group, 2016]{farasa}
Group, A. L.~T. (2016).
\newblock Farasa toolkit.

\bibitem[Habash et~al., 2007]{habash2007arabic}
Habash, N., Rambow, O., and Roth, R. (2007).
\newblock Arabic diacritization with and without a corpus.
\newblock {\em Arabian Journal for Science and Engineering}, 32(1C):3--18.

\bibitem[Hassan and Hassan, 2010]{hassan2010sakhr}
Hassan, H. and Hassan, H. (2010).
\newblock Sakhr arabic diacritization system.
\newblock {\em Proceedings of the International Conference on Language Resources and Evaluation (LREC)}, pages 3870--3876.

\bibitem[Hifny, 2021]{recent_ad}
Hifny, Y. (2021).
\newblock Recent advances in arabic syntactic diacritics restoration.
\newblock In {\em ICASSP 2021-2021 IEEE International Conference on Acoustics, Speech and Signal Processing (ICASSP)}, pages 7768--7772. IEEE.

\bibitem[Howard and Ruder, 2018]{textclassif}
Howard, J. and Ruder, S. (2018).
\newblock Universal language model fine-tuning for text classification.

\bibitem[Inoue et~al., 2021]{camelbert}
Inoue, G., Alhafni, B., Baimukan, N., Bouamor, H., and Habash, N. (2021).
\newblock The interplay of variant, size, and task type in {A}rabic pre-trained language models.
\newblock In {\em Proceedings of the Sixth Arabic Natural Language Processing Workshop}, Kyiv, Ukraine (Online). Association for Computational Linguistics.

\bibitem[Khan et~al., 2023]{education}
Khan, M., Khan, M., and Khan, M. (2023).
\newblock Natural language processing and its use in education.
\newblock {\em International Journal of Advanced Computer Science and Applications}, 5(12).

\bibitem[Maamouri et~al., 2004]{patb}
Maamouri, M., Bies, A., Buckwalter, T., and Mekki, W. (2004).
\newblock The penn arabic treebank: Building a large-scale annotated arabic corpus.
\newblock In {\em NEMLAR conference on Arabic language resources and tools}, volume~27, pages 466--467. Cairo.

\bibitem[Maamouri et~al., 2012]{arz}
Maamouri, M., Krouna, S., Tabessi, D., Hamrouni, N., and Habash, N. (2012).
\newblock Egyptian arabic morphological annotation guidelines.

\bibitem[Madhfar and Qamar, 2020]{effective_dl_ad}
Madhfar, M. A.~H. and Qamar, A.~M. (2020).
\newblock Effective deep learning models for automatic diacritization of arabic text.
\newblock {\em IEEE Access}, 9:273--288.

\bibitem[Mahdaouy et~al., 2021]{mahdaouy2021deep}
Mahdaouy, A.~E., Mekki, A.~E., Essefar, K., Mamoun, N.~E., Berrada, I., and Khoumsi, A. (2021).
\newblock Deep multi-task model for sarcasm detection and sentiment analysis in arabic language.
\newblock {\em arXiv preprint arXiv:2106.12488}.

\bibitem[Mubarak et~al., 2019a]{system_diac}
Mubarak, H., Abdelali, A., Darwish, K., Eldesouki, M., Samih, Y., and Sajjad, H. (2019a).
\newblock A system for diacritizing four varieties of arabic.
\newblock In {\em Proceedings of the 2019 Conference on Empirical Methods in Natural Language Processing and the 9th International Joint Conference on Natural Language Processing (EMNLP-IJCNLP): System Demonstrations}, pages 217--222.

\bibitem[Mubarak et~al., 2019b]{highly_eff}
Mubarak, H., Abdelali, A., Sajjad, H., Samih, Y., and Darwish, K. (2019b).
\newblock Highly effective arabic diacritization using sequence to sequence modeling.
\newblock In {\em Proceedings of the 2019 Conference of the North American Chapter of the Association for Computational Linguistics: Human Language Technologies, Volume 1 (Long and Short Papers)}, pages 2390--2395.

\bibitem[OpenAI, 2023]{gpt4}
OpenAI (2023).
\newblock Gpt-4 technical report.
\newblock {\em arXiv}.

\bibitem[Qamar, 2008]{shialibrary}
Qamar, S.~I. (2008).
\newblock Shia library (a great collection of books from shia sect).
\newblock Online Resource.

\bibitem[Qin et~al., 2021]{advers_ad}
Qin, H., Chen, G., Tian, Y., and Song, Y. (2021).
\newblock Improving arabic diacritization with regularized decoding and adversarial training.
\newblock In {\em Proceedings of the 59th Annual Meeting of the Association for Computational Linguistics and the 11th International Joint Conference on Natural Language Processing (Volume 2: Short Papers)}, pages 534--542.

\bibitem[Rashwan et~al., 2015]{RDI}
Rashwan, M. A.~A., Al~Sallab, A.~A., Raafat, H.~M., and Rafea, A. (2015).
\newblock Deep learning framework with confused sub-set resolution architecture for automatic arabic diacritization.
\newblock {\em IEEE/ACM Transactions on Audio, Speech, and Language Processing}, 23(3):505--516.

\bibitem[Reimers and Gurevych, 2019]{reimers2019sentencebert}
Reimers, N. and Gurevych, I. (2019).
\newblock Sentence-bert: Sentence embeddings using siamese bert-networks.

\bibitem[Ryding, 2005]{msa}
Ryding, K.~C. (2005).
\newblock {\em A Reference Grammar of Modern Standard Arabic}.
\newblock Cambridge University Press.

\bibitem[Taji et~al., 2014]{taji2014impact}
Taji, D., Habash, N., and Elkoury, R. (2014).
\newblock The impact of arabic diacritization on part-of-speech tagging.
\newblock In {\em Proceedings of the International Conference on Language Resources and Evaluation (LREC)}, pages 1306--1311.

\bibitem[Thompson and Alshehri, 2022]{diac_trans}
Thompson, B. and Alshehri, A. (2022).
\newblock Improving {A}rabic diacritization by learning to diacritize and translate.
\newblock In {\em Proceedings of the 19th International Conference on Spoken Language Translation (IWSLT 2022)}, pages 11--21, Dublin, Ireland (in-person and online). Association for Computational Linguistics.

\bibitem[Vaswani et~al., 2017]{attention}
Vaswani, A., Shazeer, N., Parmar, N., Uszkoreit, J., Jones, L., Gomez, A.~N., Kaiser, L., and Polosukhin, I. (2017).
\newblock Attention is all you need.
\newblock In {\em Proceedings of the 31st International Conference on Neural Information Processing Systems}, NIPS'17, page 6000–6010, Red Hook, NY, USA. Curran Associates Inc.

\bibitem[Wolf et~al., 2020]{hf_transformers}
Wolf, T., Debut, L., Sanh, V., Chaumond, J., Delangue, C., Moi, A., Cistac, P., Rault, T., Louf, R., Funtowicz, M., Davison, J., Shleifer, S., von Platen, P., Ma, C., Jernite, Y., Plu, J., Xu, C., Scao, T.~L., Gugger, S., Drame, M., Lhoest, Q., and Rush, A.~M. (2020).
\newblock Huggingface's transformers: State-of-the-art natural language processing.

\bibitem[Yousef et~al., 2019]{apcd2}
Yousef, W.~A., Ibrahime, O.~M., Madbouly, T.~M., and Mahmoud, M. (2019).
\newblock Learning meters of arabic and english poems with recurrent neural networks: a step forward for language understanding and synthesis.
\newblock {\em ArXiv}, abs/1905.05700.

\bibitem[Zaheer et~al., 2021]{zaheer2021big}
Zaheer, M., Guruganesh, G., Dubey, A., Ainslie, J., Alberti, C., Ontanon, S., Pham, P., Ravula, A., Wang, Q., Yang, L., and Ahmed, A. (2021).
\newblock Big bird: Transformers for longer sequences.

\bibitem[Zalmout and Habash, 2020]{joint_diac_lemma}
Zalmout, N. and Habash, N. (2020).
\newblock Joint diacritization, lemmatization, normalization, and fine-grained morphological tagging.
\newblock In {\em Proceedings of the 58th Annual Meeting of the Association for Computational Linguistics}, pages 8297--8307, Online. Association for Computational Linguistics.

\bibitem[Zerrouki, 2020]{mishkal}
Zerrouki, T. (2020).
\newblock {\em Towards An Open Platform For Arabic Language Processing}.
\newblock PhD thesis.

\end{thebibliography}
\end{document}